\documentclass[letterpaper]{article} %
\usepackage{aaai25}
\usepackage{times}  %
\usepackage{helvet}  %
\usepackage{courier}  %
\usepackage[hyphens]{url}  %
\usepackage{graphicx} %
\usepackage{natbib}  %
\usepackage{caption} %
\usepackage{algorithm}
\usepackage{algorithmic}

\usepackage{newfloat}
\usepackage{listings}
\DeclareCaptionStyle{ruled}{labelfont=normalfont,labelsep=colon,strut=off} %
\floatstyle{ruled}
\newfloat{listing}{tb}{lst}{}
\floatname{listing}{Listing}
\usepackage{color}
\usepackage{amsmath}
\usepackage{tikz}
\usepackage{dsfont}
\usepackage{subcaption}
\usepackage{amssymb}
\usepackage{booktabs}
\usepackage{placeins}

\title{Fairness and Sparsity within Rashomon sets:\\ Enumeration-Free Exploration and Characterization}
\author {
    Lucas Langlade\textsuperscript{\rm 1},
    Julien Ferry\textsuperscript{\rm 2},
    Gabriel Laberge\textsuperscript{\rm 3}
    Thibaut Vidal\textsuperscript{\rm 2},
}
\affiliations {
    \textsuperscript{\rm 1}Ecole nationale des ponts et chaussées, Paris, France\\
    \textsuperscript{\rm 2}CIRRELT \& SCALE-AI Chair in Data-Driven Supply Chains, Department of Mathematics and Industrial Engineering, Polytechnique Montréal, Canada\\
    \textsuperscript{\rm 3}Department of Computer Engineering and Software Engineering, Polytechnique Montréal, Canada\\
    julien.ferry@polymtl.ca,thibaut.vidal@polymtl.ca
}

\usepackage{bibentry}
\def\indicatorfunction{\mathds{1}}

\newcommand{\vect}[1]{\boldsymbol{#1}}

\def\R{\mathbb{R}}
\def\N{\mathbb{N}}

\def\nexamples{N}
\def\nfeatures{M}
\def\attributes[#1]{\vect{x}_{#1}}
\def\labels[#1]{y_{#1}}
\def\predictions[#1]{\hat{\labels[#1]}}
\def\pred[#1]{\mathbf{y^{\widehat}}_{#1}}
\def\example{i}
\def\feature{j}
\def\data{S}

\def\lossgeneric{\ell}
\def\loss{\lossgeneric_{\text{0/1}}}

\newcommand\emploss[2]{\widehat{\mathcal{L}}_{#1}(#2)}

\def\H{\mathcal{H}}
\def\h{h}

\def\Rashomon{\mathcal{R}(\H, \epsilon, S)}

\DeclareMathOperator*{\argmin}{arg\,min}

\def\coeffsparsity{C}
\def\minus{\scalebox{0.5}[1.2]{$-$}}

\def\agroup[#1]{G_{#1}}
\def\agrouppos[#1]{\agroup[#1]^+}
\def\fairnessvalue{d}
\def\valueSP{\fairnessvalue_{SP}}
\def\valueEO{\fairnessvalue_{EO}}

\def\interpretable{I}
\def\property{\phi}
\def\sparsityvalue{\alpha}

\def\slim{\texttt{SLIM}}
\def\slimcoefficient[#1]{\lambda_{#1}}
\def\slimcoefficients{\vect{\lambda}}
\def\slimcoefficientsrange{\Omega}
\def\slimbestloss{\emploss{\data}{\slimcoefficients{}_S}}
\def\dummyloss{\emploss{\data}{h_{\text{maj}}}}
\def\slimlossvariables[#1]{z_{#1}}
\def\slimpredictions[#1]{\hat{y}_{#1}}
\def\bigMone[#1]{O'_{#1}}
\def\bigMtwo[#1]{O_{#1}}
\def\slimcoeffvalueindex{\omega}
\def\slimcoeffchoicevar{u}

\def\ddsbestloss{\emploss{\data}{(\nodeused[\data], \hyperplanea[\data], \hyperplaneb[\data])_{\node \in \treestructureinternal}}}
\def\treestructure{\mathcal{V}}
\def\treestructurearcs{\mathcal{A}}
\def\graph{\mathcal{G}}
\def\treestructureinternal{\treestructure^I}
\def\treestructureleaves{\treestructure^C}
\def\node{v}
\def\nodebis{u}
\def\nodeter{w}
\def\class{c}
\def\ddsdepth{D}
\def\ddslevel{l}

\def\nodeused[#1]{d_{#1}}
\def\nodesused{\vect{d}}
\def\hyperplanea[#1]{\vect{a}_{#1}}
\def\hyperplaneb[#1]{{b}_{#1}}
\def\ddswvars[#1,#2]{w_{#1#2}}
\def\ddsflowvars{f}
\def\ddslambdavar{g}
\def\ddslinkingvars{t}

\def\updated#1{\textcolor{black}{#1}}

\begin{document}

\maketitle

\begin{abstract}
We introduce an enumeration-free method based on mathematical programming to precisely characterize various properties such as fairness or sparsity within the set of ``good models", known as \emph{Rashomon set}. This approach is generically applicable to any hypothesis class, provided that a mathematical formulation of the model learning task exists. It offers a structured framework to define the notion of business necessity and evaluate how fairness can be improved or degraded towards a specific protected group, while remaining within the Rashomon set and maintaining any desired sparsity level. 

We apply our approach to two hypothesis classes: scoring systems and decision diagrams, leveraging recent mathematical programming formulations for training such models. As seen in our experiments, the method comprehensively and certifiably quantifies tradeoffs between predictive performance, sparsity, and fairness. We observe that a wide range of fairness values are attainable, ranging from highly favorable to significantly unfavorable for a protected group, while staying within less than 1\% of the best possible training accuracy for the hypothesis class. Additionally, we observe that sparsity constraints limit these tradeoffs and may disproportionately harm specific subgroups. As we evidenced, thoroughly characterizing the tensions between these key aspects is critical for an informed and accountable selection of models.
\end{abstract}

\section{Introduction}

The increasing reliance on machine learning models for high-stakes decision-support tasks, such as predictive justice~\cite{angwin2016machine}, 
hiring~\cite{langenkamp2020hiring}, and medicine~\cite{DBLP:conf/aaai/0001CDM21} raises important ethical questions and is subject to legal requirements. 
For instance, article 13 of the recent EU AI Act\footnote{\url{https://artificialintelligenceact.eu/article/13/}} mandates transparency for AI-based systems classified 
as ``high-risk",  a category that encompasses a broad spectrum of applications. %
This legal and ethical context highlights the importance of developing predictive models that are inherently \emph{interpretable} and \emph{sparse}. %
Fairness is another critical consideration, reinforced by legal frameworks such as the ``80 percent rule'' for statistical parity~\cite{DBLP:conf/kdd/FeldmanFMSV15} established by the US Equal Employment Opportunity Commission~\cite{uniformguidelinesemployeeselection} in the context 
of hiring.%

When training a machine learning model, the primary objective is typically to maximize its utility. However, multiple models with equivalent performance can produce substantially different predictions, a phenomenon known as predictive multiplicity. This observation gave rise to the concept of $\epsilon$-Rashomon sets~\citep{breiman2001statistical,fisher2019all}, which encompass all models within a given hypothesis class whose utility deviates by no more than $\epsilon$ from the optimal value. Due to predictive 
multiplicity, models within an $\epsilon$-Rashomon set can exhibit markedly different ---sparsity or fairness--- properties.

Different methods have been proposed to explore Rashomon sets for specific hypothesis classes.
On one hand, enumeration-based 
methods~\citep{DBLP:conf/nips/XinZ0TSR22,DBLP:journals/corr/abs-2204-11285,ciaperoni2024efficient} sample all models within 
the Rashomon set of interest. This can be computationally prohibitive since the Rashomon sets might contain an untractable number of models. On the other hand, enumeration-free approaches~\citep{coker2021theory,DBLP:conf/aaai/Watson-DanielsP23,fisher2019all,zhong2024exploring,coston2021characterizing} can characterize properties across all models in the Rashomon set without explicitly enumerating them. Despite their efficiency, existing enumeration-free approaches tend to underestimate the size of Rashomon sets and their corresponding fairness-utility tradeoffs. This limitation arises from their reliance on convex upper bounds (e.g., logistic loss, hinge loss) as approximations of the actual model error ($0/1$ loss). Moreover, they do not explore the effects of sparsity requirements, which further influence these tradeoffs.

To overcome these limitations, we introduce an enumeration-free method based on mathematical programming to precisely explore various properties within 
Rashomon sets, such as fairness or sparsity. Our approach provides a structured, quantitative framework for evaluating the concept of ``business necessity''~\cite{grover1995business}, a legal argument often used by companies with unbalanced employment outcomes among protected groups. A key aspect of proving ``business necessity'' is demonstrating that no alternative employment policy could achieve the same objectives with less discriminatory impact. Within our framework, this translates to asserting that all high-performing models within a given hypothesis class exhibit a disparate impact toward a specific group.
Equivalently, the search for less discriminatory alternative models is becoming a legal requirement~\citep{DBLP:journals/corr/abs-2406-06817}, and our framework automates this task. More precisely, it can be used to precisely and provably bound the achievable fairness values within given performance and sparsity levels.
Our main contributions are: 
\begin{itemize}
    \item We propose an enumeration-free framework for exploring the properties of models within the Rashomon set, focusing on key aspects such as fairness and sparsity. This framework is broadly applicable to any hypothesis class where the learning process can be formulated as a mathematical program.
    \item To illustrate the versatility of our framework, we apply it to two different hypothesis classes: scoring systems and decision diagrams. The associated source code is
    openly accessible on our repository\footnote{\url{https://github.com/vidalt/Rashomon-Explorer}}, 
    under a MIT license. %
    \item We conduct extensive experiments validating the effectiveness of our approach in certifiably quantifying the tradeoffs (and tensions) between predictive performance, fairness, and sparsity for a given hypothesis class. Our results precisely characterize the range of fairness values achievable under specified sparsity and performance constraints. Additionally, we observe that sparsity does not come for free ---imposing stringent sparsity requirements significantly limits the achievable tradeoffs between fairness and performance.
\end{itemize} 

\section{Technical Background}
\label{sec:technical_background}

\paragraph{Supervised Machine Learning.}
Let \smash{$\data:=(\attributes[\example], \labels[\example])_{\example=1}^\nexamples$} be a dataset in which each example \smash{$\example \in \{1..\nexamples\}$} is characterized
 by a feature vector \smash{$\attributes[\example]\in\R^{\nfeatures}$} with $\nfeatures$ attributes and a binary label \smash{$\labels[\example]\in\{\minus 1,1\}$}.
The objective of a supervised learning algorithm is to produce a predictive model \smash{$\h: \R^{\nfeatures}\rightarrow\{\minus 1, 1\}$} from a given hypothesis 
space $\H$ that minimizes a loss function \smash{$\lossgeneric:\{\minus 1, 1\}\times\{\minus 1, 1\}\rightarrow \R_+$} encoding the error between a predicted and 
actual outcome. In practice, the empirical loss \smash{$\emploss{\data}{\h}:=\frac{1}{\nexamples}\sum_{\example=1}^\nexamples\lossgeneric(\h(\attributes[\example]), 
\labels[\example])$} is minimized to get a good predictor $\h_\data$:
\begin{align}
    \h_\data\in \argmin_{\h\in\H}\,\,\,\,\emploss{\data}{\h} + \text{Regularization}(\h).
    \label{eq:erm}
\end{align}
The regularization term steers an \emph{a priori} preference toward certain hypotheses in $\H$. For this study, we adopt the commonly used $0/1$ loss, defined as  \smash{$\loss(\predictions[], 
\labels[]):=\indicatorfunction [\predictions[]=\labels[]]$},
with $\predictions[]=\h(\attributes[])$.

\paragraph{Fairness.} Undesirable biases ---specifically harming some individuals or demographic groups--- can be embedded in the dataset $\data$, introduced or amplified by the learning algorithm, or arise at any other step of the machine learning pipeline~\cite{mehrabi2021survey}. Because learning such spurious correlations and using them for decision-making raises ethical questions and is often legally prohibited, different notions of fairness have been proposed~\cite{10.1145/3194770.3194776}. In particular, \emph{statistical fairness} metrics are widely adopted due to their quantifiability and their ability to align with legal standards, such as the ``80 percent rule" for statistical parity~\cite{DBLP:conf/kdd/FeldmanFMSV15}, as outlined by the US Equal Employment Opportunity Commission~\cite{uniformguidelinesemployeeselection} in the context of hiring practices.\\
Statistical fairness metrics assess disparities in specific statistical measures between different \emph{protected groups}, defined by the values of \emph{sensitive features} (e.g., race, gender). The goal of these metrics is to ensure that such features do not influence individual outcomes. Typically, these measures are derived from the confusion matrix of the predictor $\h$.
Let \smash{$\agroup[1] \subset \data$} and \smash{$\agroup[2] \subset \data$} represent two protected groups differentiated by a given sensitive feature. For example, in a hiring context, \smash{$\agroup[1]$} might represent the set of male applicants, while \smash{$\agroup[2]$} represents the set of female applicants. We consider two widely used fairness metrics:

\textbullet\ \emph{Statistical parity}~\citep{dwork2012fairness} quantifies the difference in positive prediction rates (e.g., acceptance rates for job applicants) between the two protected groups:
    \begin{align}
        \valueSP(\h,\data) := \frac{\sum_{\example\in \agroup[1]}  \indicatorfunction [\h(\attributes[\example])=1]}{|\agroup[1]|} - 
    \frac{\sum_{\example \in \agroup[2]} \indicatorfunction [\h(\attributes[\example])=1]}{|\agroup[2]|} \label{eq:sp}
    \end{align}
\textbullet\ \emph{Equal opportunity}~\citep{hardt2016equality} measures the difference in true positive rates (e.g., acceptance rates for genuinely qualified applicants) between the two protected groups:
    \begin{align}
        \valueEO(\h,\data) := \frac{\sum_{\example\in \agrouppos[1]}  \indicatorfunction [\h(\attributes[\example])=1]}{|\agrouppos[1]|} - 
    \frac{\sum_{\example \in \agrouppos[2]} \indicatorfunction [\h(\attributes[\example])=1]}{|\agrouppos[2]|} \label{eq:eo}
    \end{align}
with \smash{$\agrouppos[1]=\{\example \in \agroup[1] | \labels[\example] = 1\}$} and \smash{$\agrouppos[2]=\{\example \in \agroup[2] | \labels[\example] = 1\}$}.
For both metrics, values closer to $0$ indicate better fairness in the model. Positive values suggest a bias favoring \smash{$\agroup[1]$}, while negative values indicate a bias toward \smash{$\agroup[2]$}.

\paragraph{Interpretability.}
It can be defined as ``the ability to explain or to present something in understandable terms to a human"~\cite{doshi2017towards}. It is a critical property for ensuring the trustworthiness of machine learning systems and is often a legal requirement in real-world applications. One possible approach to achieving interpretability is through \emph{post-hoc} explanations~\cite{guidotti2018survey} of black-box models, which aim to clarify either individual decisions or the model's overall behavior.
However, such methods can be unreliable in certain contexts and are vulnerable to manipulation~\cite{aivodji2019fairwashing,slack2020fooling}. 
An alternative is to develop inherently \emph{interpretable} models, such as decision trees or rule lists, which do not share these weaknesses~\cite{rudin2019stop}. While interpretability lacks a universal definition, \emph{sparsity} 
(such as the number of nodes in a decision tree) is often used as a proxy~\cite{rudin2022interpretable}.
Enforcing sparsity constraints effectively restricts the hypothesis space to a more interpretable subset, $\H_{\interpretable} \subset \H$~\citep{dziugaite2020enforcing}.

\paragraph{Mathematical Programming.} 
Mathematical programming involves defining a set of \emph{decision variables}, each constrained to a specific domain, and specifying \emph{constraints} that describe relationships between these variables. A general-purpose solver is used to find \emph{feasible} assignments of the decision variables that satisfy the given constraints. When an \emph{objective function} is provided, the solver seeks a feasible solution that either maximizes or minimizes the function. The types of domains and constraints that can be expressed depend on the chosen paradigm. For example, \emph{Mixed-Integer Linear Programming} (MILP) solvers can accommodate both continuous and discrete variables but are restricted to linear constraints and objective functions.

\section{Related Works: Exploring Rashomon Sets}
\label{sec:related_works}

Beyond predictive performance, other model properties, such as fairness and sparsity, are often desirable. Since these properties are typically not aligned with maximizing predictive performance, it is necessary to tolerate a small drop in performance, quantified as \smash{$\epsilon$}, to search for alternative models \smash{$\h_\text{alt}$} satisfying \smash{$\emploss{\data}{\h_{\text{alt}}} \leq \emploss{\data}{\h_{\data}} + \epsilon$}.  The set of such alternative models is referred to as the \emph{Rashomon set}~\citep{breiman2001statistical,fisher2019all}, defined as:
\begin{align}
    \Rashomon := \{\h\in \H : \emploss{\data}{\h}\leq \emploss{\data}{\h_{\data}}+\epsilon\}.
\end{align}
Rashomon sets have been studied in the context of \emph{predictive multiplicity}, demonstrating that 
different models can have conflicting predictions on a substantial subset of data~\citep{marx2020predictive,DBLP:conf/nips/HsuC22,DBLP:conf/aaai/Watson-DanielsP23}. 
A key result from \citet{marx2020predictive} establishes that for any alternative $\h_{\text{alt}}\in \Rashomon$, the following tight bound holds:
\smash{$\frac{1}{\nexamples}\sum_{\example=1}^\nexamples \indicatorfunction[\h_{\text{alt}}(\attributes[\example])\neq \h_{\data}(\attributes[\example])]\leq 2\emploss{\data}
{\h_{\data}}+\epsilon$}.
This implies that even with a small $\epsilon$, models within the Rashomon set can differ significantly in their predictions whenever the empirical loss \smash{$\emploss{\data}{\h_{\data}}$} is non-zero. For instance, if \smash{$\h_{\data}$} has an empirical loss of 10\%, alternative models in the Rashomon set could disagree with \smash{$\h_{\data}$} on up to 20\% of the dataset. Similarly, for a predictor \smash{$\h_{\data}$} with a 25\% empirical loss, disagreements with alternative models could extend to 50\% of the dataset. These substantial differences emphasize that models within the Rashomon set, despite achieving nearly equivalent predictive performance, can vary markedly in their predictions. This variability has important implications for fairness, as fairness metrics such as statistical parity (Equation~(\ref{eq:sp})) and equal opportunity (Equation~(\ref{eq:eo})) are aggregates of predictions across demographic subgroups. Consequently, the Rashomon set may contain alternative models with more desirable fairness or sparsity properties. However, identifying such models efficiently remains a significant challenge. Existing methods for exploring the Rashomon set can be categorized into two main approaches: \emph{enumeration-based} and \emph{enumeration-free}.

Enumeration-based methods sample all models within the Rashomon set (or a substantial subset thereof). Existing approaches have applied enumeration to the Rashomon sets of rule lists~\citep{DBLP:journals/corr/abs-2204-11285}, rule sets~\citep{ciaperoni2024efficient}, and decision trees~\citep{DBLP:conf/nips/XinZ0TSR22} using branch-and-bound techniques. These methods explore the combinatorial hypothesis space $\H$ while leveraging error lower bounds to prune the search space efficiently. These past works have shown that competing models exhibit different fairness properties. However, a key limitation of these methods is their reliance on enumerating (and storing) a large number of models. \updated{While methods were proposed to sample only a subset of models from the Rashomon set -- e.g., through bootstrapping~\citep{DBLP:conf/aaai/CooperLCBSGK0Z24} or shuffling~\citep{DBLP:conf/fat/GaneshC0S23} --  they do not aim at exploring the whole Rashomon set and cannot provide bounds on the extrema of a functional.}

\begin{table*}[t]
    \begin{tabular}{llll}
    \toprule
    Source   & Hypothesize class $\H$     & Loss $\ell$             & Functional $\phi$ \\
    \midrule
    \citet{coker2021theory}  & Linear & hinge loss   & prediction\\
    \citet{DBLP:conf/aaai/Watson-DanielsP23}  & Linear   & logistic loss & prediction \\
    \citet{fisher2019all} & Linear/Kernels  & hinge loss  & feature importance \\
    \citet{zhong2024exploring} & Additive (GAM) & logistic loss  & feature importance \\
    \citet{coston2021characterizing} & Linear  & logistic loss & fairness metrics \\
    \textbf{Ours}   & \textbf{Linear/Decision-Diagrams} & \textbf{0-1 loss}  & \textbf{fairness metrics}\\
    \bottomrule
    \end{tabular}
    \caption{Summary of enumeration-free Rashomon set exploration methods for binary classification tasks.}\label{tab:rashomon}
\end{table*}

Alternatively, enumeration-free methods focus on the identification of models within the Rashomon set that achieve the extreme values of a specific functional \smash{$\property: \H \rightarrow \R$}.
This approach allows targeted exploration of the Rashomon set by optimizing for particular properties without exhaustive enumeration. Previous work has investigated the min-max range of the functional $\property(\h)=\h(\attributes[])$ for 
linear models with the hinge-loss \citep{coker2021theory} and the logistic loss \citep{DBLP:conf/aaai/Watson-DanielsP23}. 
Other studies have explored the extreme values of feature importance scores for linear models under the squared or hinge loss \citep{fisher2019all}, or Generalized Additive Models (GAMs) under the logistic loss \citep{zhong2024exploring}. 
Finally, the min-max range of the functional $\property$ underlying fairness metrics (cf. Equations (\ref{eq:sp}) \& (\ref{eq:eo})) has been characterized for linear models with logistic loss \citep{coston2021characterizing}.
Table~\ref{tab:rashomon} summarizes these previous works.%

\updated{Considering the ``true" $0/1$ loss is desirable, since it exactly quantifies the utility of the trained model (the proportion of individuals whose outcome was incorrectly predicted). However, it makes the problem more difficult to solve since it is not continuous nor convex. Using convex upper bounds such as logistic or hinge losses makes the problem more tractable, but leads to underestimating the Rashomon set, hence arbitrarily limiting the tradeoffs between other desiderata (e.g., fairness and sparsity). In turn, discrete optimization tools (such as mathematical programming) are appropriate to directly handle the 0-1 loss and exactly characterize the Rashomon set. They are also particularly well-suited to learn models that are inherently interpretable (which is the focus of this paper), such as rule-based or tree-based ones -- which intrinsically have a combinatorial structure. Based on these observations, our study:}
\begin{itemize}
    \item explores the range of unfairness values within the Rashomon sets using the ``true'' $0/1$ loss, whereas previous works relied on convex upper bounds; %
    \item characterizes the effect of sparsity constraints on the range of possible disparities within the Rashomon set;
    \item provides a framework applicable to many hypothesis classes $\H$
    provided their learning process can be formulated as a mathematical optimization problem.
\end{itemize}

\updated{We now discuss two closely related works that also explore the fairness properties of model classes, though from different perspectives. Considering models within all possible input-output mappings for a finite dataset, \citet{dai2025intentional} define the notion of \emph{largest possible Rashomon Set}, and propose methods to efficiently bound the achievable fairness values within them. While these bounds hold for any hypothesis class, there might not exist any model within a given hypothesis class implementing the chosen input-output mapping, hence they can be arbitrarily loose. On the contrary, our work aims at providing tight bounds for a given hypothesis class, along with their associated models for a chosen sparsity level. \citet{DBLP:conf/fat/SimsonPK24} rather characterize the set of fairness values reachable depending on the design of the ML pipeline (e.g., preprocessing and evaluation choices), while we precisely bound fairness for all possible models within the Rashomon set for a fixed dataset and hypothesis class.}

\section{Exploring Rashomon Sets Through Mathematical Programming}

We first introduce our generic framework for exploring the Rashomon set of a given hypothesis class whose learning is formulated as a mathematical program. We then instantiate it for two widely used classes of interpretable models, namely scoring systems and decision diagrams.

\subsection{Generic Framework}

As stated in Equation~(\ref{eq:erm}), the goal of a machine learning algorithm is to explore the hypothesis space $\H$ to identify a model $\h_\data$ 
that minimizes (on a training dataset $\data$) a given objective function, which consists of its empirical loss \smash{$\emploss{\data}{\h_\data}$} and, optionally, a regularization 
term. We focus on the common scenario where the regularization term measures the model's sparsity, with the tradeoff between sparsity and 
predictive performance governed by a coefficient $\coeffsparsity$. The general mathematical formulation of this learning process is:
\begin{align}
    \min_{\h\in \H}&\,\,\,\,\emploss{\data}{\h} + \coeffsparsity \cdot \text{Sparsity}(\h). \label{eq:opt_general}
\end{align}
The model's structure and parameters are encoded through \emph{decision variables}, while its internal predictions and adherence to the hypothesis 
space are enforced through a set of \emph{constraints}. 

The objective of our proposed framework is to provably determine the maximum and minimum values of a given fairness metric (along with the 
corresponding models) subject to a desired sparsity constraint while remaining within an $\epsilon$-Rashomon set of the hypothesis space $\H$. To achieve this, we first solve Problem~(\ref{eq:opt_general}) with $\coeffsparsity=0$ to obtain the optimal empirical loss value 
\smash{$\emploss{\data}{\h_{\data}}$}, which by definition constitutes the reference value for the Rashomon set computation. We then formulate and solve the following problem:
\begin{align}
    \min_{\h}&\,\,\,\,\valueSP(\h,\data)  \label{eq:opt_fairness}\\
    \text{s.t.}& \quad \h \in \H \nonumber\\
    & \quad \text{Sparsity}(\h) \leq \sparsityvalue \nonumber\\
    & \quad \emploss{\data}{\h}\leq \emploss{\data}{\h_{\data}}+\epsilon \nonumber
\end{align}
Here, $\sparsityvalue$ represents the desired sparsity, which sets an upper bound on the model's size, and \smash{$\valueSP(\h,\data)$} is the 
statistical parity metric, although any other fairness measure can replace it. By reversing the sign of the objective, the full range of fairness 
values within the $\epsilon$-Rashomon set can be characterized. Additionally, by varying the desired sparsity level~$\sparsityvalue$, the impact 
of sparsity constraints on the accuracy-fairness tradeoff can be further explored.

\subsection{Instantiation for Scoring Systems}

Scoring systems are sparse linear classification models with integer coefficients, widely used in fields like medicine and criminal justice due to their interpretability~\citep{rudin2022interpretable}. To make a prediction with such a model on a given example $\attributes[]$, one multiplies each feature's value $x_j$ by its corresponding coefficient $\slimcoefficient[j]$ selected within an acceptable range of values $\slimcoefficientsrange{}_\feature \subset \N$, sums the results, and compares the total to a fixed threshold. %
The hypothesis space of scoring systems %
is then:
\begin{align*}
     &\H := \big\{\attributes[] \mapsto\  \text{sign}(\attributes[]^T\slimcoefficients{})
    \mid \slimcoefficient[j]\in \slimcoefficientsrange{}_\feature, \feature=1..\nfeatures\big\}.
\end{align*}
Table~\ref{tab:example_scoring_system} presents an example scoring system trained on the Default of Credit Card Clients dataset~\cite{yeh2009comparisons}.
The classification task involves predicting whether a person will default on payment based on demographic information and payment histories. In addition to the coefficients associated with the $\nfeatures$ features (only non-zero ones are shown), an additional \textbf{threshold} is included. This threshold is usually handled by concatenating an additional feature with a value of $1$ to all examples before training or inference.
As visible in the table, the model’s interpretability allows for straightforward identification of the features influencing predictions. For instance, features indicating delays in previous payments or high payment amounts are associated with an increased likelihood of predicting a default on the next payment. However, the model also exhibits a bias against females, as the attribute ``SEX\_Female'' increases the computed score, thereby increasing the probability of predicting a default for females.
This aligns with the measured statistical parity value (Equation~(\ref{eq:sp})) of $-0.046$, whose negativity indicates a higher positive prediction rate for group $\agroup[2]$ (females) over group $\agroup[1]$. In this example, interpretability facilitates the detection of such discriminations.

\slim{} \emph{(Supersparse Linear Integer Model)}~\cite{slim2014} is a MILP formulation designed to learn optimal 
scoring systems. We use it within our framework to instantiate the learning problem defined in Problem~(\ref{eq:opt_general}). The original \slim{} formulation aims at finding the coefficients $\slimcoefficients{}_\data$ minimizing the following objective:
    \begin{align}
        \min_{\slimcoefficients{}} \,\,\, & \sum_{\example = 1}^{\nexamples} 
        \indicatorfunction[\labels[\example] \attributes[\example]^T \slimcoefficients{} \leq 0] + 
        \coeffsparsity \| \slimcoefficients{} \| _0 
        \label{eq:slim_obj}
    \end{align} %
where $\slimcoefficients{}$ is the vector of coefficients within the scoring system, $\attributes[\example]^T \slimcoefficients{}$ is the scoring system's total score 
for example~$\example$ (whose sign determines the output label), and $\coeffsparsity$ is a regularization coefficient. Then, \smash{$\sum_{\example = 1}^{\nexamples} 
\indicatorfunction[\labels[\example] \attributes[\example]^T \slimcoefficients{} \leq 0]$} computes the $0/1$ empirical loss of the model, while 
\smash{$\| \slimcoefficients{} \| _0$} is a sparsity regularizer, penalizing the number of non-zero coefficients. %

We now present our modified formulation, which instantiates Problem~(\ref{eq:opt_fairness}) to characterize fairness and sparsity within the Rashomon set of scoring systems. Recall that the optimal loss value \smash{$\slimbestloss{}$} is first obtained by solving the original \slim{} formulation with \smash{$\coeffsparsity=0$} (i.e., ensuring that objective~(\ref{eq:slim_obj}) focuses solely on predictive performance).

\begin{align}
\min_{\slimcoefficients{}} \quad & \frac{\sum_{\example \in \agroup[1]} \slimpredictions[\example]}{|\agroup[1]|} - 
    \frac{\sum_{\example \in \agroup[2]} \slimpredictions[\example]}{|\agroup[2]|}\label{constr:slimfair_objective}\\
\text{s.t.} \quad & \slimcoefficient[\feature] = \sum_{\slimcoeffvalueindex{} \in \slimcoefficientsrange_\feature} 
    \slimcoeffvalueindex{} \cdot \slimcoeffchoicevar_{j \slimcoeffvalueindex{}} &\feature \in \{1,..,\nfeatures\} \label{constr:slimfair_coeff_set}\\
 & \sum_{\slimcoeffvalueindex{} \in \slimcoefficientsrange_\feature}
    \slimcoeffchoicevar_{\feature \slimcoeffvalueindex{}} \leq 1 &\feature \in \{1,..,\nfeatures\}\label{constr:slimfair_coeff_choice}\\
 & \sum_{\feature=1}^\nfeatures \sum_{\slimcoeffvalueindex{} = 1}^{\slimcoefficientsrange_\feature} \slimcoeffchoicevar_{\feature \slimcoeffvalueindex{}} 
    \leq \sparsityvalue &\text{(Sparsity)}\label{constr:slimfair_sparsity}\\
 & \frac{1}{\nexamples} \sum_{\example = 1}^{\nexamples} \slimlossvariables[\example] \leq \slimbestloss+\epsilon &\text{(Performance)}\label{constr:slimfair_rashomon_set}\\
 & \bigMone[\example] \slimlossvariables[\example] \geq \gamma - \labels[\example] \attributes[\example]^T\slimcoefficients{} & 
    \example \in \{1,..,\nexamples\}\label{constr:slimfair_loss_1}\\
 & \bigMtwo[\example](1 - \slimlossvariables[\example]) \geq \labels[\example] \attributes[\example]^T\slimcoefficients{} & 
    \example \in \{1,..,\nexamples\}\label{constr:slimfair_loss_2}\\
 & \slimpredictions[\example] = (1-\slimlossvariables[\example])\indicatorfunction[\labels[\example] = 1] \nonumber \\ 
 & \quad \quad + \slimlossvariables[\example] \indicatorfunction[\labels[\example] = -1] & \example \in \{1,..,\nexamples\}\label{constr:slimfair_predictions}\\
 & \slimcoefficient[\feature] \in \slimcoefficientsrange{}_\feature  &\feature \in \{1,..,\nfeatures\} \nonumber\\
 & \slimlossvariables[\example] \in \{ 0, 1 \} &\example \in \{1,..,\nexamples\} \nonumber\\
 & \slimpredictions[\example] \in \{ 0, 1 \} &\example \in \{1,..,\nexamples\} \nonumber \\
 &  \slimcoeffchoicevar_{\feature \slimcoeffvalueindex{}} \in \{ 0, 1 \} &\hspace{-100pt}\feature \in \{1,..,\nfeatures\},~\slimcoeffvalueindex{} \in \slimcoefficientsrange_\feature\nonumber 
\end{align}

Each coefficient $\slimcoefficient[\feature]$ associated to feature $\feature$ within the scoring system must take a value within a user-defined domain~$\slimcoefficientsrange{}_\feature$.
 Specifically, Constraint~(\ref{constr:slimfair_coeff_set}) ensures that the coefficient $\slimcoefficient[\feature]$ takes value $\slimcoeffvalueindex{} \in \slimcoefficientsrange_\feature$ if and only if $\slimcoeffchoicevar_{\feature \slimcoeffvalueindex{}}=1$. Constraint~(\ref{constr:slimfair_coeff_choice}) guarantees that at most one value $\slimcoeffvalueindex{} \in \slimcoefficientsrange_\feature$ is set to $1$.
Note that $\slimcoefficient[\feature]=0$ if none of the variables $\slimcoeffchoicevar_{\feature \slimcoeffvalueindex{}}$ equals $1$.

\begin{table}[t!]
\centering
\begin{tabular}{|cc|}
\hline
\multicolumn{1}{|c|}{\textbf{Feature}}           & \textbf{Coefficient}      \\ \hline
\multicolumn{1}{|c|}{EDUCATION\_University}       & 2                         \\ \hline
\multicolumn{1}{|c|}{PAY\_0\_Pay\_delay$\geq$1}   & 5                         \\ \hline
\multicolumn{1}{|c|}{PAY\_2\_Pay\_delay$\geq$1}   & 5                         \\ \hline
\multicolumn{1}{|c|}{PAY\_6\_Pay\_delay$\geq$1}   & 2                         \\ \hline
\multicolumn{1}{|c|}{PAY\_AMT6\_high}             & 2                         \\ \hline
\multicolumn{1}{|c|}{SEX\_Female}                 & 2                         \\ \hline
\multicolumn{1}{|c|}{\textbf{Threshold}}                  & \textbf{-10}                       \\ \hline
\multicolumn{2}{|c|}{\textit{\textbf{Predict $+1$ if total is $>0,$ $-1$ otherwise}}} \\ \hline
\end{tabular}
\caption{Example scoring system trained on the Default of Credit Card Clients dataset, belonging to the $20\%$-Rashomon set, exhibiting $0.842$ training accuracy and $0.80$ test accuracy, as well as $-0.046$ training statistical parity.}\label{tab:example_scoring_system}
\end{table}

Objective~(\ref{constr:slimfair_objective}) represents the statistical parity metric introduced in Equation~(\ref{eq:sp}). By minimizing it, we aim to find the scoring system that maximally favors the protected group $\agroup[2]$ over $\agroup[1]$. Reversing the sign of this difference allows us to optimize the fairness value in the opposite direction. Constraint~(\ref{constr:slimfair_rashomon_set}) limits the hypothesis space to the $\epsilon$-Rashomon set, 
leveraging the previously computed optimal loss \smash{$\slimbestloss$} (as defined in Equation~(\ref{eq:slim_obj})). Constraint~(\ref{constr:slimfair_sparsity}) restricts the number of non-zero coefficients in $\slimcoefficients{}$ to at most $\sparsityvalue$, thereby enforcing sparsity.

The remaining constraints handle the intermediate computations of the scoring system's predictive performance and predictions.
Specifically, the loss variables $\slimlossvariables[]$ indicate whether each example $\example$ is incorrectly classified: 
\smash{$\slimlossvariables[\example] = \indicatorfunction[\labels[\example] \attributes[\example]^T \slimcoefficients{} \leq 0]$}. These variables are determined by  
Constraints~(\ref{constr:slimfair_loss_1}--\ref{constr:slimfair_loss_2}), which compare the sign of each example $\example$'s predictions 
\smash{$(\attributes[\example]^T\slimcoefficients{})$} with its true label $\labels[\example]$. Note that \smash{$\bigMone[\example \in \{1..\nexamples\}]$} and 
\smash{$\bigMtwo[\example \in \{1..\nexamples\}]$} are pre-computed constants large enough to enforce the constraints, and $\gamma$ is a small constant 
representing a margin, ensuring that \smash{$(\labels[\example] \attributes[\example]^T\slimcoefficients{})$} for all examples $\example$ is lower-bounded. 

In the original \slim{} formulation, because the sum of the loss variables was minimized in the objective, 
Constraint~(\ref{constr:slimfair_loss_1}) alone was sufficient to set $\slimlossvariables[\example]$ to $1$ if and only if example $\example$ is misclassified. As this is no longer the case here, we must additionally include Constraint~(\ref{constr:slimfair_loss_2}) to force $\slimlossvariables[\example]$ to $0$ in case of correct classification.

Finally, the predictions $\slimpredictions[]$ are computed leveraging the loss variables and the actual labels (given as input constants to the model) 
through Constraint~(\ref{constr:slimfair_predictions}). For each example $\example$, we then have: 
$\slimpredictions[\example] = \indicatorfunction[\attributes[\example]^T \slimcoefficients{} > 0]$. 

This formulation precisely determines the extent to which each protected group can be favored over the other, given a specified sparsity level $\sparsityvalue$ 
(maximum number of non-zero coefficients) and predictive performance threshold (defined by the $\epsilon$-Rashomon set). By varying $\sparsityvalue$ and $\epsilon$, 
one can explore the tradeoffs between these different desiderata.

\subsection{Instantiation for Decision Diagrams}

Decision diagrams are popular interpretable models exhibiting a top-down hierarchical structure similar to trees. Yet, unlike decision trees, their branches can be merged. This fundamental property avoids the replication and fragmentation problems of decision trees~\citep{oliver1992decision,kohavi1994bottom,ddiagrams}, hence enhancing interpretability. Formally, a decision diagram is a rooted directed acyclic graph $\graph=(\treestructure,\treestructurearcs)$, where each internal node $\node \in \treestructureinternal$ represents a splitting hyperplane and each terminal node $\node \in \treestructureleaves$ is uniquely associated with a prediction $\class_{\node}$. This hypothesis class generalizes rule-lists (\smash{$\H_{\text{rule-list}}\subset \H_{\text{diagrams}}$}), 
so investigating its $\epsilon$-Rashomon set is an enumeration-free alternative to the approach of \citet{DBLP:journals/corr/abs-2204-11285}.
As with \slim{}, the objective \smash{$\valueSP(\h,\data)$}, and Sparsity/Performance constraints of Problem~(\ref{eq:opt_fairness}) are easily 
expressed as linear functions of decision variables, allowing for a MILP formulation.

We build upon the original MILP formulation by \citet{ddiagrams} for learning optimal decision diagrams for classification. In essence, given a user-specified maximum structure, the formulation aims to determine which nodes and edges should be utilized within this structure and how their splitting hyperplanes should be defined. Sparsity is then computed as the number 
$\sparsityvalue$ of active (utilized) internal nodes. The objective is as follows:
\begin{align}
    \min_{(\nodeused[], \hyperplanea[], \hyperplaneb[])_{\node \in \treestructureinternal}} \,\,\, & \sum_{\example = 1}^{\nexamples} 
        \slimlossvariables[\example] + 
        \coeffsparsity \| \nodesused{} \| _0. 
        \label{eq:dds_objective}
\end{align}
Here, for each example $\example$, the loss variable $\slimlossvariables[\example]$ indicates whether it is misclassified, so $\smash{\sum_{\example = 1}^{\nexamples} \slimlossvariables[\example]}$ computes the $0/1$ loss. For each internal node within the predefined structure $\node \in \treestructureinternal$, the variable $\nodeused[\node] \in \{0, 1\}$ indicates whether it is used in the final structure. The term $\| \nodesused{} \| _0$ quantifies the sparsity of the resulting decision diagram by counting the number of nodes used in the trained structure.
Finally, for each internal node $\node$ where $\nodeused[\node]=1$, variables $(\hyperplanea[\node],\hyperplaneb[\node])$ define the hyperplane corresponding to the multivariate split performed by this node.
This objective effectively instantiates Problem~(\ref{eq:opt_general}).

\updated{We hereafter provide our modified formulation, which instantiates Problem~(\ref{eq:opt_fairness}) to characterize fairness and sparsity within the Rashomon set of decision diagrams. Recall that the optimal loss value \smash{$\ddsbestloss{}$} is first obtained by solving the original MILP formulation with \smash{$\coeffsparsity=0$} (i.e., ensuring that Objective~(\ref{eq:dds_objective}) focuses solely on predictive performance). The following objective represents the statistical parity metric introduced in Equation~(\ref{eq:sp}):
\begin{flalign}\min_{(\nodeused[], \hyperplanea[], \hyperplaneb[])_{\node \in \treestructureinternal}} \quad \frac{\sum_{\example \in \agroup[1]} \slimpredictions[\example]}{|\agroup[1]|} - 
    \frac{\sum_{\example \in \agroup[2]} \slimpredictions[\example]}{|\agroup[2]|}\label{constr:dds_objective}
    \end{flalign}
By minimizing it, we aim to find the decision diagram that maximally favors the protected group $\agroup[2]$ over $\agroup[1]$ in terms of positive prediction rate. Reversing the sign of this difference allows us to constrain the fairness value in the opposite direction. We hereafter introduce the different constraints that must be satisfied while optimizing Objective~(\ref{constr:dds_objective}).
Constraint~(\ref{constr:dds_sparsity}) restricts the number of active nodes in $\nodesused{}$ 
to at most $\sparsityvalue$, thereby enforcing sparsity.
Constraint~(\ref{constr:dds_rashomon_set}) limits the hypothesis space to the $\epsilon$-Rashomon set, utilizing the previously computed optimal loss \smash{$\ddsbestloss{}$}:
\begin{flalign}
&\sum_{\node \in \treestructureinternal} \nodeused[\node] \leq \sparsityvalue 
&\text{(Sparsity)}\label{constr:dds_sparsity}\\
& \frac{1}{\nexamples} \sum_{\example = 1}^{\nexamples} \slimlossvariables[\example] \leq \ddsbestloss{}+\epsilon \hspace{-100pt} &\text{(Performance)}\label{constr:dds_rashomon_set}
\end{flalign}
The $0/1$ loss associated to each example $\example \in \{1,..,\nexamples\}$ is then computed as follows:
\begin{flalign}
& \slimlossvariables[\example] = \sum_{\node \in \treestructureleaves} \indicatorfunction[\labels[\example]\neq\class_{\node}] \ddswvars[\example,\node] && \label{constr:dds_loss}\\
& \slimpredictions[\example] = (1-\slimlossvariables[\example])\indicatorfunction[\labels[\example] = 1] + \slimlossvariables[\example] \indicatorfunction[\labels[\example] = -1] && \label{constr:dds_predictions}
\end{flalign}
Constraint~(\ref{constr:dds_loss}) sets the loss variable $\slimlossvariables[\example]=1$ if and only if example $\example$ is assigned to a terminal node $\node \in \treestructureleaves$ whose predicted class $\class_{\node}$ differs from the example's true label $\labels[\example]$. Constraint~(\ref{constr:dds_predictions}) then uses the loss variables $\slimlossvariables[]$ to determine the decision diagram's predictions~$\slimpredictions[]$.
The remaining constraints are unchanged and model the structure of the constructed decision diagram. Below, we briefly discuss the role of each constraint, and we refer to \citet{ddiagrams} for a more comprehensive explanation. For each example $\example \in \{1,..,\nexamples\}$, Constraints~(\ref{constr:dds_flow_1}--\ref{constr:dds_flow_3}) model the flow of each example through the nodes of the decision diagram: 
\begin{flalign}
& \ddswvars[\example,\node]^+ + \ddswvars[\example,\node]^- = \begin{cases}
    1~\text{if}~\node = 0 \\
\sum_{\nodebis \in \delta^-(\node)}(\ddsflowvars^+_{\example\nodebis\node} + \ddsflowvars^-_{\example\nodebis\node})
\end{cases} &\node \in \treestructureinternal \label{constr:dds_flow_1} \\
& \ddswvars[\example,\nodebis]^- = \sum_{\node\in\delta^+(\nodebis)} \ddsflowvars^-_{\example\nodebis\node} &\nodebis\in \treestructureinternal \label{constr:dds_flow_2}\\
& \ddswvars[\example,\nodebis]^+ = \sum_{\node\in\delta^+(\nodebis)} \ddsflowvars^+_{\example\nodebis\node} &\nodebis\in \treestructureinternal \label{constr:dds_flow_3}
\end{flalign}
\vspace{-\baselineskip}
\begin{flalign}
& \sum_{\nodebis \in \treestructureinternal_\ddslevel} \ddswvars[\example,\nodebis]^- \leq 1 - \ddslambdavar_{\example\ddslevel} &\ddslevel \in \{0,..,\ddsdepth-1\}\label{constr:dds_flow_integrity_1}\\
& \sum_{\nodebis \in \treestructureinternal_\ddslevel} \ddswvars[\example,\nodebis]^+ \leq \ddslambdavar_{\example\ddslevel} &\ddslevel \in \{0,..,\ddsdepth-1\}\label{constr:dds_flow_integrity_2}
\end{flalign}
Specifically, $\delta^-(\nodebis)$ (respectively, $\delta^+(\nodebis)$) represents the set of possible predecessors (respectively, successors) of node $\nodebis$ in the user-provided decision diagram structure.
The variable $\ddswvars[\example,\nodebis]^-$ (respectively, $\ddswvars[\example,\nodebis]^+$) takes a non-zero value when example $\example$ passes through node $\nodebis$ on the negative (respectively, positive) side of the separating hyperplane. Additionally, the variable $\ddsflowvars^-_{\example\nodebis\node}$ (respectively, $\ddsflowvars^+_{\example\nodebis\node}$) models the flow from the negative (respectively, positive) side of $\nodebis$ to other nodes~$\node$.  Constraints~\mbox{(\ref{constr:dds_flow_integrity_1}--\ref{constr:dds_flow_integrity_2})} ensure flow integrity using the binary variable $\ddslambdavar_{\example\ddslevel}$, which determines, for each example~$\example$, whether it follows the negative or positive side at each level $\ddslevel \in \{0,..,\ddsdepth-1\}$, $\ddsdepth$ being the depth of the decision diagram.
For each node $\nodebis \in \treestructureinternal$, Constraints~(\ref{constr:dds_set_node_used_1}--\ref{constr:dds_set_node_used_2}) specify that it is used in the decision diagram ($\nodeused[\nodebis]=1$) if and only if it is connected to or from another node: 
\begin{flalign}
& \nodeused[\nodebis] = \sum_{\node \in \delta^+(\nodebis)} \ddslinkingvars^+_{\nodebis\node} = \sum_{\node \in \delta^+(\nodebis)} \ddslinkingvars^-_{\nodebis\node} &  \label{constr:dds_set_node_used_1}\\
& \nodeused[\node] \leq \sum_{\nodebis \in \delta^-(\node)}(\ddslinkingvars^+_{\nodebis\node} + \ddslinkingvars^-_{\nodebis\node}) &\node \in \treestructureinternal\setminus\{0\} \label{constr:dds_set_node_unused}\\
& \ddslinkingvars^+_{\nodebis\node} + \ddslinkingvars^-_{\nodebis\node} \leq \nodeused[\node] &\node\in\delta^+(\nodebis)\label{constr:dds_set_node_used_2}
\end{flalign}
The binary variable $\ddslinkingvars^-_{\nodebis\node}$ (respectively, $\ddslinkingvars^+_{\nodebis\node}$) indicates that node $\nodebis \in \treestructureinternal$ links to node $\node$ on the negative (respectively, positive) side. Note that both the root and terminal nodes are excluded from these constraints, as they are always used. Constraints~(\ref{constr:link_flow_1}--\ref{constr:link_flow_2}) connect the linking variables to the examples' flows, for each node $\nodebis \in \treestructureinternal$:
\begin{flalign}
& \ddsflowvars^+_{\example\nodebis\node} \leq \ddslinkingvars^+_{\nodebis\node} &\node\in\delta^+(\nodebis),~\example \in \{1,..,\nexamples\} \label{constr:link_flow_1}\\
& \ddsflowvars^-_{\example\nodebis\node} \leq \ddslinkingvars^-_{\nodebis\node} &\node\in\delta^+(\nodebis),~\example \in \{1,..,\nexamples\} \label{constr:link_flow_2}
\end{flalign}
Constraints~(\ref{constr:symmetry_breaking_1}--\ref{constr:symmetry_breaking_2}) implement symmetry breaking, as many equivalent topologies could result from the previously defined constraints and variables:
\begin{flalign}
& \ddslinkingvars^-_{\nodebis\node} + \sum_{\nodeter \in \delta^+(\nodebis), \nodeter\leq \node} \ddslinkingvars^+_{\nodebis\nodeter} \leq 1 &\nodebis\in \treestructureinternal,~\node\in\delta^+(\nodebis) \label{constr:symmetry_breaking_1}
\end{flalign}
For each level $\ddslevel \in \{2,..,\ddsdepth-1\}$, a weak in-degree ordering is enforced for each pair of nodes $\nodebis,\node\in\treestructureinternal_\ddslevel$ such that $\nodebis<\node$:
\begin{flalign}
& \sum_{\nodeter\in \delta^-(\nodebis)}(\ddslinkingvars^+_{\nodeter\nodebis} +\ddslinkingvars^-_{\nodeter\nodebis}) \geq \sum_{\nodeter\in \delta^-(\node)}(\ddslinkingvars^+_{\nodeter\node} +\ddslinkingvars^-_{\nodeter\node})  \label{constr:symmetry_breaking_2}
\end{flalign}
Finally, for each example $\example \in \{1,..,\nexamples\}$, Constraints~(\ref{constr:hyperplane_1}--\ref{constr:hyperplane_2}) ensure consistency between the hyperplane variables and the example's flow, while Constraint~(\ref{constr:leaves_assignments}) determines the terminal node $\node$ to which it is assigned by setting $\ddswvars[\example,\node]$ based on the previously computed flows:
\begin{flalign}
& (\ddswvars[\example,\node]^- = 1) \implies (\hyperplanea[\node]^T\attributes[\example] + \gamma \leq \hyperplaneb[\node]) &\node \in \treestructureinternal \label{constr:hyperplane_1}\\
& (\ddswvars[\example,\node]^+ = 1) \implies (\hyperplanea[\node]^T\attributes[\example] > \hyperplaneb[\node]) &\node \in \treestructureinternal \label{constr:hyperplane_2}\\
& \ddswvars[\example,\node] = \sum_{\nodebis\in\delta^-(\node)} (\ddsflowvars^+_{\example\nodebis\node} + \ddsflowvars^-_{\example\nodebis\node}) &\node\in \treestructureleaves \label{constr:leaves_assignments}
\end{flalign}
}

\section{Experimental Study}

Our numerical experiments serve two main objectives.
First, we demonstrate the applicability and effectiveness of our framework in characterizing fairness and sparsity within Rashomon sets, through an instantiation for two hypothesis classes.
Second, we explore the interplays between the three desiderata, highlighting the main trends.

\subsection{Experimental Setup}\label{sec:setup}

\paragraph{Datasets.} We consider three datasets widely used in the fair and interpretable machine learning literature. First, the UCI Adult Income dataset~\cite{Dua:2019} contains records on $32,561$ individuals from the 1994 U.S. census, described by $36$ binary attributes. The classification task is to predict whether an individual earns more than \$50K per year. In our experiments, $\agroup[1]$ represents males and $\agroup[2]$ represents females.
Second, the Default of Credit Card Clients dataset~\cite{yeh2009comparisons} includes demographic information and payment histories for $29,986$ individuals in Taiwan, each described by $21$ attributes. The task is to predict whether a person will default on payment, with $\agroup[1]$ as males and $\agroup[2]$ as females.
Third, the COMPAS dataset~\cite{angwin2016machine} contains data on $7,214$ criminal offenders in Broward County, Florida, described by $27$ binary attributes. The task is to predict whether an individual will re-offend within two years. Here, $\agroup[1]$ represents African-Americans, and $\agroup[2]$ represents the rest of the population.

\paragraph{Learning Procedure.}  For each dataset, we randomly sub-sample training sets $\data$ of size $\nexamples=500$, with the remaining 
examples used as a test set, as this permits a fast and unbiased evaluation based on optimal solutions of the underlying mathematical models. \updated{Note that we also report some results using larger training set sizes $\nexamples$ in the Appendix~\ref{appendix:scalability} to illustrate the scalability of our method and the consistency of our empirical findings.}
We generate five different random splits and report both the average values and standard deviations 
in our experiments. The two fairness metrics considered are statistical parity (Equation~(\ref{eq:sp})) and equal opportunity 
(Equation~(\ref{eq:eo})).
For each random split of each dataset, we determine the optimal loss \smash{$\emploss{\data}{\h_{\data}}$} and the majority classifier 
loss \smash{$\dummyloss{}$}. Then, the $\epsilon$ parameter is chosen so that the loss upper bound lies between these two
extreme losses : \smash{$(1-p)\emploss{\data}{\h_{\data}} + p\dummyloss{}$} with \smash{$p \in \{ 1\%, 5\%, 10\%, 20\% \}$}. 
Notably, the $0\%$-Rashomon set includes the optimal models and the $100\%$-Rashomon set includes all models 
not worse than a majority classifier. \updated{The predictive performances of the reference models (achieving the optimal loss \smash{$\emploss{\data}{\h_{\data}}$}) are reported in the Appendix~\ref{appendix:complete_results}. They confirm that the models' training accuracies are in line with the literature and that they generalize well.}

\paragraph{Hyperparameters.} 
Our experiments using scoring systems use the same set of possible values for all coefficients: %
$\slimcoefficientsrange{}_{\feature \in \{1,..,\nfeatures\}} = 
\{0, \pm1, \pm2, \pm5, \pm10, \pm20, \pm30, \pm50 \}$. %
Sparsity values $\sparsityvalue$ (i.e., maximum numbers of non-zero 
coefficients in the scoring systems) range from $1$ to $\nfeatures+1$ (to account for the additional bias coefficient).
Based on preliminary experiments, we fix the skeleton of the decision diagrams to a maximum of 12 internal nodes, distributed across $5$ consecutive levels as follows: $(1,2,3,3,3)$. We consider sparsity values (i.e., the maximum number of active nodes in the decision diagrams) ranging from $\sparsityvalue=4$ to $\sparsityvalue=12$.

\begin{figure*}[t!]
  \centering
   \includegraphics[width=0.45\linewidth]{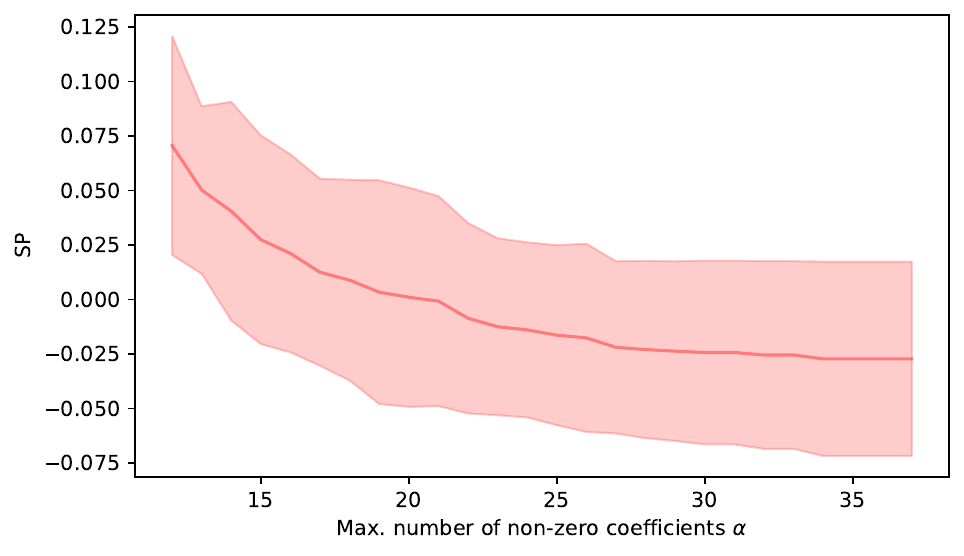}
    \hfill
   \includegraphics[width=0.45\linewidth]{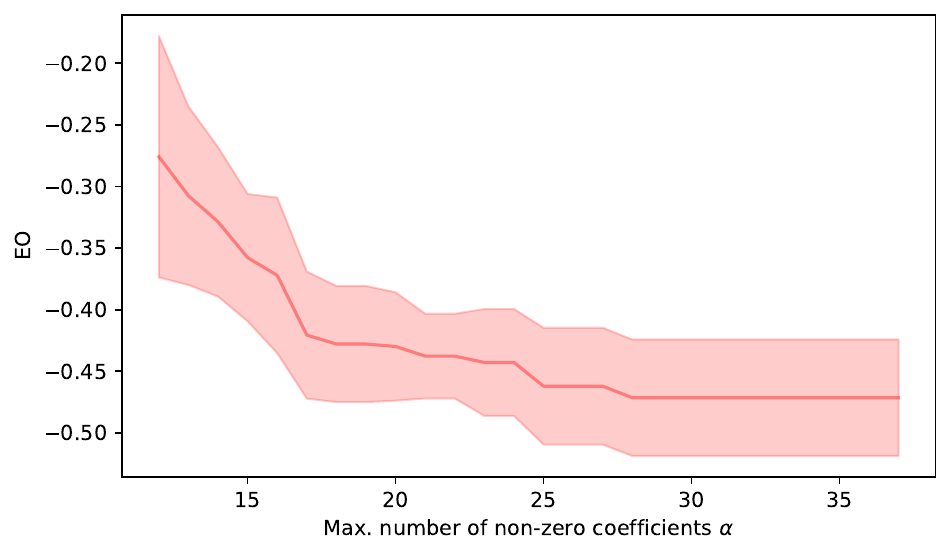}
\caption{Minimum statistical parity (SP) and equal opportunity (EO) achievable on the UCI Adult Income dataset, within a $20\%$-Rashomon set for scoring systems, as a function of the chosen sparsity value $\sparsityvalue$. Negative values favour group $\agroup[2]$ (females) over group $\agroup[1]$ (males), for predicting a high salary. We display both average value and standard deviation.}
\label{fig:effect1}
\end{figure*}

\paragraph{Exploration of the Rashomon set.}
 We use the \texttt{Gurobi} solver~\citep{gurobi} through its Python binding to solve Problems~(\ref{constr:slimfair_objective}) for scoring systems and~(\ref{constr:dds_objective}) for decision diagrams. Each solver execution is done on 16 threads using a computing cluster with Intel Platinum 8260 Cascade Lake @2.4GHz CPUs.
 To speed up our experiments, we exploit the fact that increasing either the allowed sparsity value $\sparsityvalue$ or the Rashomon set parameter $\epsilon$ relaxes the problem, so we can rely on previously found solutions to hot start the solver. Specifically, each run (for a fixed dataset, random split, sparsity value $\sparsityvalue$, and Rashomon set parameter $\epsilon$) is limited to one hour of CPU time and 36 GB of RAM. For runs where no feasible solution was found or optimality was not proven, we reuse solutions obtained from more constrained versions of the problem (i.e., tighter values of $\sparsityvalue$ or $\epsilon$) and restart the solver. Convergence was reached in all runs after at most five such iterations.

\subsection{Results}\label{sec:results}

We now highlight our key empirical findings and illustrate each of them with a subset of representative results. Complete results, including all datasets, fairness metrics, $\alpha$ and $\epsilon$ parameters, and larger training set sizes are provided in 
the Appendix~\ref{appendix:complete_results} for both considered hypothesis classes. %

\paragraph{Result 1. Sparsity restricts the range of achievable fairness values and may harm certain protected groups.}
As discussed earlier, tightening the enforced sparsity value $\sparsityvalue$ confines the search to a subset of the hypothesis space $\H_I \subset \H$, which can limit the tradeoffs between various objectives~\citep{dziugaite2020enforcing}, including fairness and predictive performance. While this result could be expected, our framework allows us to precisely and certifiably quantify this effect. Furthermore, the extent to which sparsity restricts the possible tradeoffs between fairness and predictive performance indicates the severity of the tension between the three desiderata.

For instance, Figure~\ref{fig:effect1} shows the minimum achievable fairness values within a $20\%$-Rashomon set of scoring systems as a function of the enforced sparsity $\sparsityvalue$ for both the statistical parity (left) and equal opportunity (right) metrics on the UCI Adult Income dataset. Negative values for both metrics indicate a bias in favor of group $\agroup[2]$ (females) in predicting high salaries. By quantifying the minimum achievable value, we effectively measure the extent to which females can be advantaged over males given the specified sparsity and performance desiderata.

As expected, tightening the sparsity $\sparsityvalue$ reduces the range of achievable fairness. This suggests that enforcing sparsity excludes models with extreme fairness values, highlighting a conflict between these two criteria. Notably, the left plot shows that scoring systems with fewer than $\sparsityvalue=20$ non-zero coefficients systematically disadvantage group $\agroup[2]$ (females), as indicated by the positive minimum fairness values. Hence, if high sparsity is legally required, the resulting outcome imbalance favoring group $\agroup[1]$ (males) could be justified under the principle of ``business necessity''. 

\paragraph{Result 2: Different fairness metrics exhibit different tradeoffs with sparsity.}
A comparison of the two plots in Figure~\ref{fig:effect1} reveals that the impact of sparsity on the minimum achievable fairness within a $20\%$-Rashomon set varies depending on the fairness metric considered. Specifically, sparsity consistently disadvantages females in terms of statistical parity (left plot). However, this is not the case for equal opportunity (right plot), where the minimum achievable value remains negative. This difference can be attributed to the fact that, as shown in Equation~(\ref{eq:eo}), equal opportunity is conditioned on the true labels and therefore aligns more closely with predictive accuracy, whereas statistical parity does not. 

\begin{table*}[htb!]
\begin{center}
\begin{subtable}{0.48\textwidth}
    \centering
    \begin{tabular}{ccccc}
            \hline
            $\alpha$ = 15 & $\epsilon$ = 1$\%$ & $\epsilon$ = 5$\%$ & $\epsilon$ = 10$\%$ & $\epsilon$ = 20$\%$
            \\ 
            \hline
            Min SP & $-0.160$ & -0.185 & -0.215 & -0.236 \\
            & $\pm 0.060$ & $\pm 0.062$ & $\pm 0.057$ & $\pm 0.048$\\ 
            \hline
            Max SP & $0.059$ & $0.077$ & $0.120$ & $0.137$\\ 
            & $\pm 0.099$ & $\pm 0.094$ & $\pm 0.077$ & $\pm 0.072$\\ 
            \hline
            \hline
            $\alpha$ = 9 & $\epsilon$ = 1$\%$ & $\epsilon$ = 5$\%$ & $\epsilon$ = 10$\%$ & $\epsilon$ = 20$\%$\\ 
            \hline
            Min SP & N/A & $-0.098$ & $-0.124$ & $-0.176$\\
            & & $\pm 0.068$ & $\pm 0.063$ & $\pm 0.067$ \\ 
            \hline
            Max SP & N/A & $-0.017$ & $-0.009$ & $0.053$ \\
            & & $\pm 0.110$ & $\pm 0.108$ & $\pm 0.057$  \\ 
            \hline
    \end{tabular} 
    \caption{Scoring systems}
    \label{tab:effect2_ss}
\end{subtable}
\hfill
\begin{subtable}{0.48\textwidth}
    \centering
    \begin{tabular}{ccccc}
            \hline
            $\alpha$ = 7 & $\epsilon$ = 1$\%$ & $\epsilon$ = 5$\%$ & $\epsilon$ = 10$\%$ & $\epsilon$ = 20$\%$
            \\ 
            \hline
            Min SP & -0.279 & -0.280 & -0.310 & -0.330 \\
            & $\pm 0.139$ & $\pm 0.138$ & $\pm 0.112$ & $\pm 0.103$\\ 
            \hline
            Max SP & 0.193 & 0.200 & 0.229 & 0.244\\ 
            & $\pm 0.090$ & $\pm 0.087$ & $\pm 0.084$ & $\pm 0.079$\\ 
            \hline
            \hline
            $\alpha$ = 4 & $\epsilon$ = 1$\%$ & $\epsilon$ = 5$\%$ & $\epsilon$ = 10$\%$ & $\epsilon$ = 20$\%$\\ 
            \hline
            Min SP & -0.272 & -0.273 & -0.292 & -0.312\\
            & $\pm 0.143$ & $\pm 0.142$ & $\pm 0.128$ & $\pm 0.114$ \\ 
            \hline
            Max SP & 0.181 & 0.197 & 0.202 & 0.215\\ 
            & $\pm 0.096$ & $\pm 0.088$ & $\pm 0.081$ & $\pm 0.073$\\ 
            \hline
    \end{tabular} 
    \caption{Decision diagrams}
    \label{tab:effect2_dds}
\end{subtable}
\caption{Minimal and maximal statistical parity (SP) achievable on the Default of Credit Card Clients dataset, within different $\epsilon$-Rashomon Sets, for two 
different sparsity values $\sparsityvalue$, for our experiments on the two considered hypothesis classes. Negative values indicate higher default in payment prediction 
rates for group $\agroup[2]$ (females) compared to group $\agroup[1]$ (males). N/A indicates that there exists no scoring system satisfying both the sparsity and 
predictive performance desiderata. We report both the average value and standard deviation.}\label{tab:effect2}
\end{center}
\end{table*}

\begin{figure*}[!ht]
    \centering
    \includegraphics[width=0.45\linewidth]
    {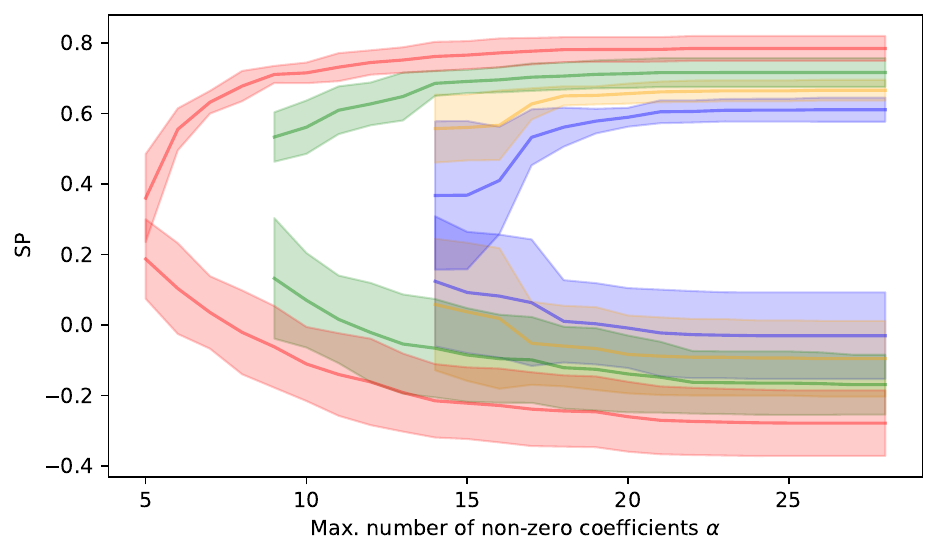}
    \hfill 
    \includegraphics[width=0.45\linewidth]
    {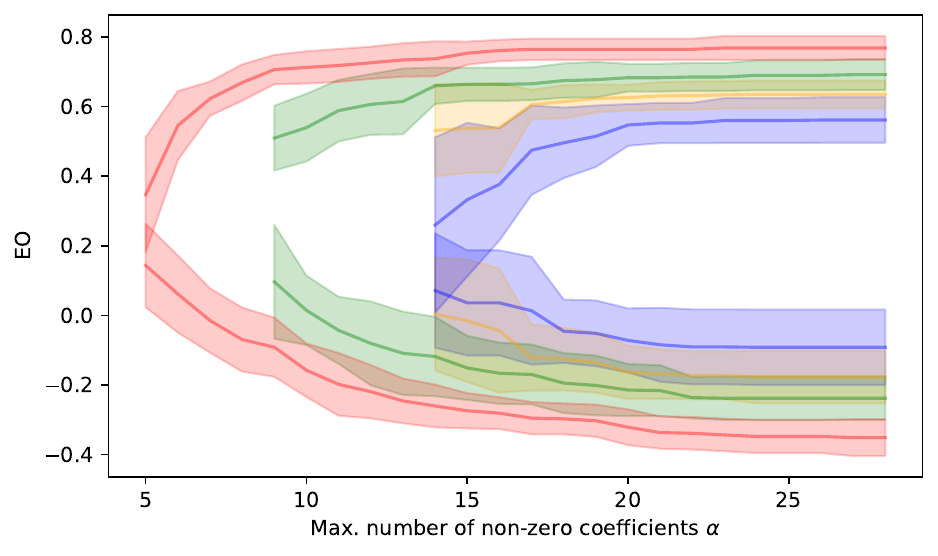}
    \includegraphics[width=0.44\linewidth]{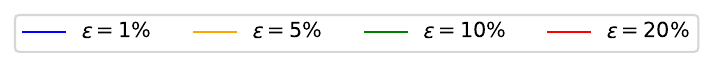}
    \caption{Minimum and maximum statistical parity (SP) and equal opportunity (EO) achievable on the COMPAS dataset, within different 
    $\epsilon$-Rashomon sets for scoring systems, as a function of the chosen sparsity value $\sparsityvalue$. Positive values indicate 
    higher recidivism prediction rates for group $\agroup[1]$ (African-Americans) compared to group $\agroup[2]$ (the remaining of the population).  
    We report both the average value (line) and standard deviation (colored area).}
\label{fig:effect3}
\end{figure*}

\paragraph{Result 3. High predictive performance requirements restrict the range of achievable fairness values.}
Table~\ref{tab:effect2_ss} shows the minimum and maximum achievable statistical parity for different Rashomon set parameters $\epsilon$ for scoring systems on the Default of Credit Card Clients 
dataset. We compare two sparsity levels: $\sparsityvalue=15$ (corresponding to the scoring system with the best achievable loss) and $\sparsityvalue=9$ (a sparser, 
arbitrary value). As previously noted, the range of achievable fairness values narrows with tighter sparsity (smaller $\sparsityvalue$).
At fixed sparsity, tightening the predictive performance constraint $\epsilon$ further restricts the achievable fairness range. Again, since enforcing tighter performance requirements amounts to shrinking the Rashomon set, this result could be expected. However, the extent to which it is the case indicates the severity of the tension between the two desiderata. Furthermore, it also allows discovering systematic biases, which, since the approach certifiably finds the minimum and achievable fairness values, can be used as legal arguments. For instance, tightening the predictive performance constraint can systematically disadvantage 
certain protected groups, as evidenced by the fact that the maximum achievable fairness becomes negative for $\epsilon \leq 10\%$ when $\sparsityvalue=9$. 
This implies that females (group $\agroup[2]$) are (on average) predicted to default on payment more often than males (group $\agroup[1]$) in a systematic manner. In other words, if one wants to build a scoring system no further than $10\%$ from the best achievable predictive performance, and with no more than $9$ non-zero coefficients (for interpretability purposes), discriminating females (in terms of statistical parity) is certifiably inevitable considering the Default of Credit Card Clients dataset.

\paragraph{Result 4. Accuracy, fairness, and sparsity have complex interplays.}
Figure~\ref{fig:effect3} plots the minimum and maximum achievable fairness values as a function of the desired sparsity level $\sparsityvalue$ for different $\epsilon$ 
parameters. The experiments were conducted on scoring systems using the COMPAS dataset and two fairness metrics. This visualization reveals the complex interplays 
between the three desiderata: predictive performance, fairness, and sparsity. As previously noted, enforcing tighter sparsity (smaller $\sparsityvalue$) narrows the range 
of achievable fairness values (represented by the gap between the minimum and maximum plotted curves of a given color). Considering tight predictive performance constraints also limits the achievable sparsity values, as indicated by the fact that the curves corresponding to small Rashomon set parameters are unable to reach the smallest sparsity values. For instance, scoring systems within the $1\%$-Rashomon set exhibit at least $14$ non-zero coefficients, while the $20\%$-Rashomon set contains scoring systems with only $5$ non-zero coefficients. Again, our approach offers a precise quantification of the tension between interpretability and predictive accuracy for a given hypothesis class. Here, ``business necessity" could justify the inability to reach a target sparsity value.

\updated{Moreover, the fact that the curves are not centered around zero highlights inherent trade-offs between predictive performance and fairness. Specifically, it means that, for a given sparsity level ($\sparsityvalue$) and predictive performance constraint ($\epsilon$), the extent to which one protected group can be favored over the other is greater than the reverse — revealing an asymmetry in how fairness can be achieved.}
Additionally, when a curve crosses the x-axis (\updated{i.e., the horizontal line at SP=0, or EO=0)}, one protected group becomes systematically 
disadvantaged across all models in the Rashomon set. For instance, scoring systems with $\sparsityvalue=5$ non-zero coefficients within a $20\%$-Rashomon set 
have a minimum statistical parity value of $18.7\%$, meaning that all models in the Rashomon set consider higher recidivism risks for African-Americans. \updated{This bias is also observed with the equal opportunity fairness metric, which is conditioned on the true labels -- suggesting that tight enough sparsity requirements systematically amplify existing biases for this experiment.}
Furthermore, as can be seen in the complete results provided in the Appendix~\ref{appendix:complete_results}, the tradeoffs between accuracy, fairness and sparsity are also influenced by the training data. \updated{In particular, the extent to which sparsity or accuracy restrain the range of achievable fairness values differs across datasets. 
While further investigations on this aspect could be conducted, key factors include the data intrinsic biases towards the considered protected groups, the respective correlations of the different attributes with the labels and the sensitive attributes, as well as the complexity of the underlying classification task.}

\paragraph{Result 5. The complexity of the hypothesis class at hand strongly influences the observed tradeoffs.}
Table~\ref{tab:effect2_dds} reports the minimum and maximum achievable statistical parity for different Rashomon set parameters $\epsilon$, based on our experiments with decision diagrams. We compare two sparsity levels: $\sparsityvalue=7$ (corresponding to the decision diagram with the best achievable loss) and $\sparsityvalue=4$ (a sparser value).
The main trends align with the key findings from our experiments on scoring systems: for a fixed sparsity $\sparsityvalue$, tightening the predictive performance requirement (smaller $\epsilon$) restricts the possible fairness ranges.
Similarly, for a fixed $\epsilon$, enforcing tighter sparsity further reduces the achievable fairness ranges.
The influence of the hypothesis class on the tradeoffs between accuracy, fairness, and sparsity is evident when comparing the results in Table~\ref{tab:effect2_dds} with those in Table~\ref{tab:effect2_ss} (which correspond to scoring systems learned on the same data splits). Decision diagrams offer a broader range of tradeoffs, with fairness ranges that are less constrained by performance and sparsity requirements. Notably, the minimum and maximum values of statistical parity systematically cross zero, implying that disparate impacts are hardly excusable by ``business necessity''. In other words, the resulting Rashomon sets systematically contain both models favoring group $\agroup[2]$ (females) and models favoring group $\agroup[1]$ (males).  These wider ranges are possible because the hypothesis class of decision diagrams is significantly more complex than that of scoring systems. Indeed, the considered decision diagrams partition the input space using multivariate splits~\citep{ddiagrams}, with each internal node functioning as a linear classifier.
In contrast, an entire scoring system corresponds to a linear classifier with integer coefficients: a single internal node of a multivariate decision diagram generalizes it, and we have: \smash{$\H_{\text{scoring systems}}\subset \H_{\text{diagrams}}$}, even for decision diagrams involving a single internal node.
However, this increased complexity comes at the expense of interpretability: understanding the resulting models is more difficult for humans due to the use of multivariate splits.
This explains why scoring systems remain very popular in high-stakes applications such as medicine~\citep{rudin2022interpretable}.
Indeed, the choice of the hypothesis space is another crucial dimension of the complex interplays between the considered ethical desiderata in machine learning. Thorough quantification of the tradeoffs between fairness, sparsity, and predictive accuracy ---facilitated by our proposed framework--- can empower stakeholders to make informed decisions when navigating these complex interdependencies.

\updated{
\paragraph{Result 6. Increasing the number of training examples $\nexamples$ can tighten the range of achievable fairness values, potentially amplifying existing discrimination.} In the Appendix~\ref{appendix:scalability}, we report results of our experiments using larger training set sizes $\nexamples$. They demonstrate that the proposed approach scales well, and that the observed trends generalize to larger values of $\nexamples$. In fact, they are even exacerbated, as the range of achievable fairness values for a given sparsity level ($\sparsityvalue$) and predictive performance threshold ($\epsilon$) narrows with increasing $\nexamples$. For instance, when $\nexamples=500$, the minimum achievable EO within $20$\%-Rashomon sets for scoring systems on the COMPAS dataset becomes positive when the number of non-zero coefficients is small ($\sparsityvalue \leq 6$), as can be seen in Figure~\ref{fig:effect3}. This means that for tight enough sparsity requirements, all models in the Rashomon set consider higher recidivism risks for African-Americans, as evidenced through a systematically higher true positive rate. When $\nexamples = 4000$ (Figure~\ref{fig:scaled_4000} in the Appendix~\ref{appendix:scalability}), this bias persists across all sparsity requirements and becomes more pronounced as sparsity is tightened.
}

\section{Discussion}

This study has demonstrated that mathematical programming approaches can be used to explore the Rashomon set of any hypothesis class without enumeration by making generic modifications to a given baseline learning problem. Specifically, we introduced a framework to characterize fairness and sparsity within the Rashomon set and validated its versatility using two popular types of interpretable models: scoring systems and decision diagrams. The resulting tools enable the identification of sparser, less discriminatory alternative models, representing a significant step toward meeting legal and ethical requirements, despite the inherent challenges~\citep{laufer2024fundamental}. \updated{The proposed methodology and software, which will be publicly released, can be used by practitioners willing to enforce fairness desiderata to estimate how much accuracy or sparsity they should be ready to sacrifice for different hypothesis classes.}

Our extensive experiments highlighted the complex interplays between predictive accuracy, fairness, and sparsity. Our framework not only certifiably quantifies these interplays but also identifies model parameters leading to extreme values, effectively guiding the search for fairer and sparser alternatives. Importantly, we observed that imposing strict predictive performance or sparsity criteria might inherently disadvantage a protected group, underscoring the need for a thorough characterization of these tradeoffs.

The research directions stemming from this work are diverse. First, we propose extending our generic framework to other hypothesis classes, such as rule-based models and tree ensembles, by leveraging recent advances in mathematical programming formulations for interpretable machine learning~\citep{DBLP:journals/eor/GambellaGN21,rudin2022interpretable}. Additionally, the declarative nature of the framework supports the integration of various additional desiderata, including alternative fairness or robustness metrics, as well as business-specific requirements. Overall, this makes it a promising tool for characterizing the tensions among key properties related to trustworthiness in machine learning.

\appendix

\bibliography{aaai25}

\clearpage

\section{Additional Experiments on Scalability}\label{appendix:scalability}

\updated{To illustrate the scalability of our method and the consistency of our findings, we present results from experiments with varying training set sizes. We focus on a representative setting: training scoring systems on the COMPAS dataset to optimize the Equal Opportunity (EO) metric within 20\%-Rashomon sets. As with our other experiments, the source code and detailed result files are provided on our repository\footnote{\url{https://github.com/vidalt/Rashomon-Explorer}}, under a MIT license.}

\updated{The experimental procedure is as described in Section~\ref{sec:setup}, with varying numbers of training examples $\nexamples \in \{500, 2000, 3000, 4000\}$. Table~\ref{tab:ssperfs_scalability} displays the predictive performances (training and test accuracies) and sparsities of the reference models (achieving the optimal loss \smash{$\emploss{\data}{\h_{\data}}$}). These results indicate that as the number of training examples $\nexamples$ increases, more non-zero coefficients are required to fit the data, slightly degrading sparsity. The simultaneous mild drop in training accuracy and rise in test accuracy suggest that scoring systems generalize slightly better with more training data, as expected.}

\begin{figure*}[htb!]
    \centering
    \begin{subfigure}{0.48\textwidth}
       \includegraphics[width=1\linewidth]
        {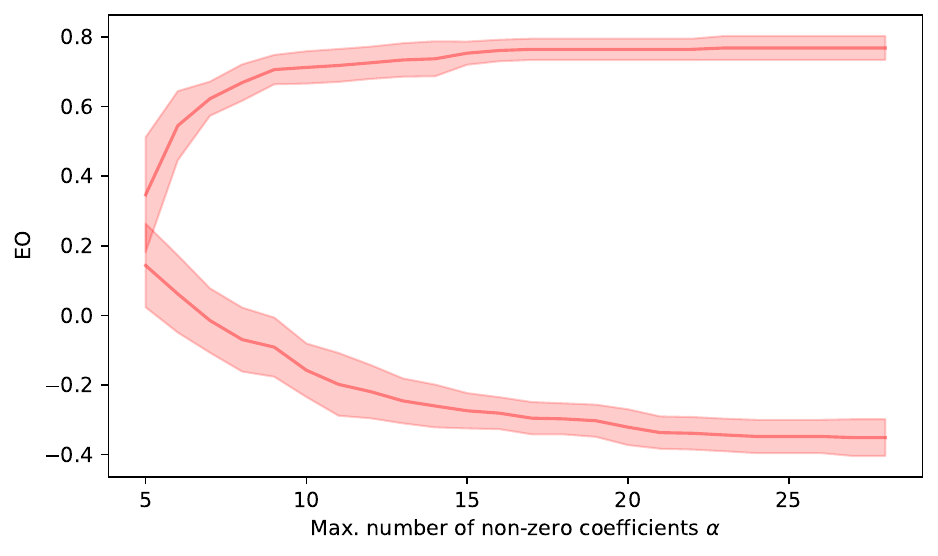}
        \caption{$\nexamples=500$}\label{fig:scaled_500}
    \end{subfigure}
    \begin{subfigure}{0.48\textwidth}
        \includegraphics[width=1\linewidth]{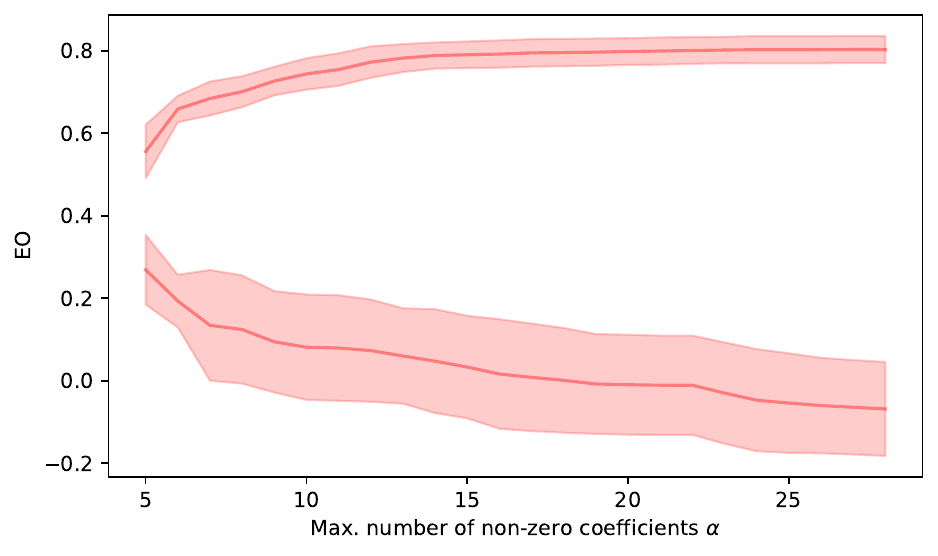}
        \subcaption{$\nexamples=2000$}\label{fig:scaled_2000}
    \end{subfigure}
    \hspace{5pt}
        \begin{subfigure}{0.48\textwidth}
       \includegraphics[width=1\linewidth]
        {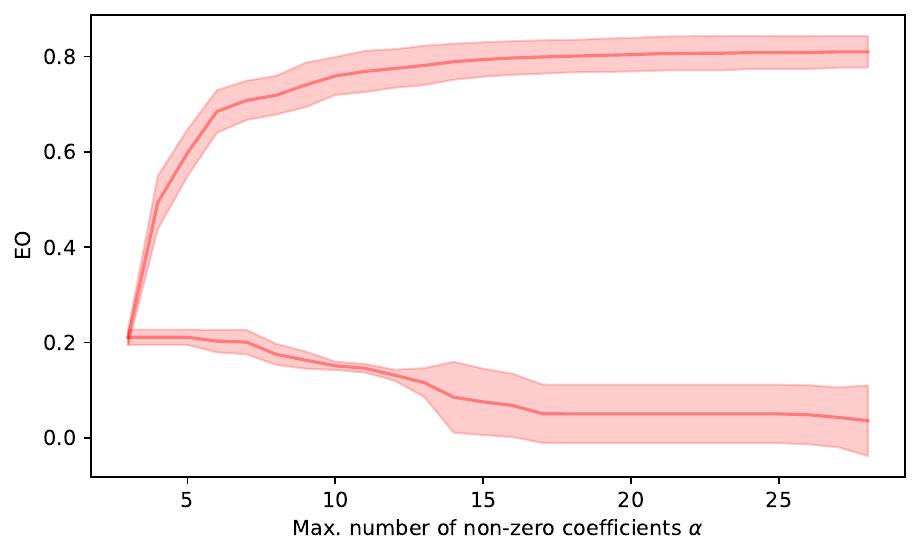}
        \caption{$\nexamples=3000$}\label{fig:scaled_3000}
    \end{subfigure}
    \begin{subfigure}{0.48\textwidth}
        \includegraphics[width=1\linewidth]{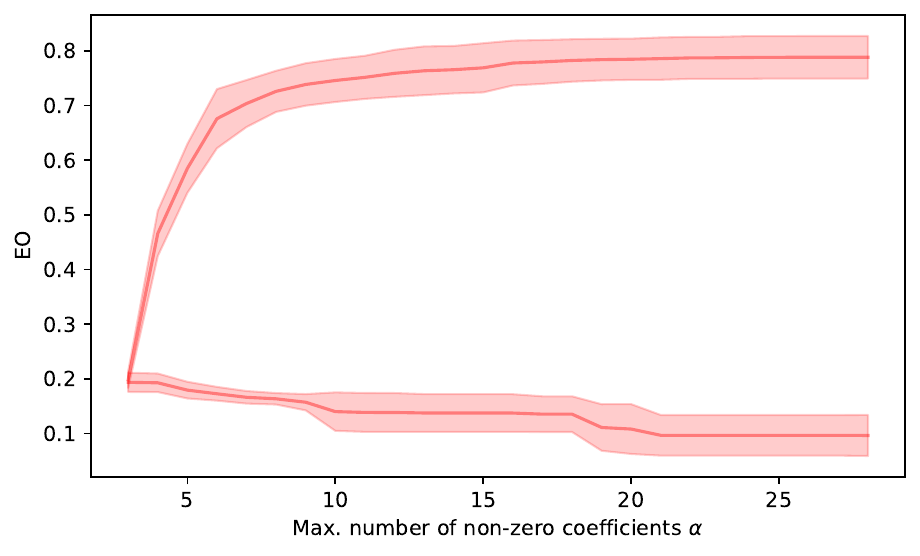}
        \subcaption{$\nexamples=4000$}\label{fig:scaled_4000}
    \end{subfigure}
    \caption{Minimum and maximum equal opportunity (EO) achievable on the COMPAS dataset, within $20\%$-Rashomon sets for scoring systems, as a function of the chosen sparsity value $\sparsityvalue$ (number of non-zero coefficients in the built scoring systems) for different numbers of training examples $\nexamples$.
    We report both the average value (line) and standard deviation (colored area).}
    \label{fig:scaled}
\end{figure*}

\updated{In Figure \ref{fig:scaled}, we plot the minimum and maximum achievable EO values (Equation~(\ref{eq:eo})) on the COMPAS dataset, within $20$\%-Rashomon sets for scoring systems, as a function of the desired sparsity level $\alpha$, for different training set sizes $\nexamples$. 
The results demonstrate that the proposed approach for thoroughly exploring Rashomon sets scales to larger training sets and that the trends observed in Section~\ref{sec:results} generalize accordingly. 
As previously noted, tightening the sparsity requirement restricts the range of achievable fairness (EO) values. We observed that for $\nexamples=500$ training examples (Figure~\ref{fig:scaled_500}, or Figure~\ref{fig:effect3} in Section~\ref{sec:results}), the minimum achievable EO within $20$\%-Rashomon sets for scoring systems becomes positive when the number of non-zero coefficients is small ($\sparsityvalue \leq 6$). This means that for tight enough sparsity requirements, all models in the Rashomon set consider higher recidivism risks for African-Americans, as evidenced through a systematically higher true positive rate. 
This trend is verified and becomes even more pronounced as the training set size $\nexamples$ increases, as shown in Figures~\ref{fig:scaled_2000}, \ref{fig:scaled_3000}, and \ref{fig:scaled_4000}. Specifically, for $\nexamples = 2000$ (Figure~\ref{fig:scaled_2000}), we observe a systematic bias disproportionately affecting group $\agroup[1]$ (African-Americans) for sparsity values $\sparsityvalue \leq 18$. When $\nexamples = 3000$ or $\nexamples = 4000$ (Figures~\ref{fig:scaled_3000} and \ref{fig:scaled_4000}), this bias persists across all sparsity requirements. Interestingly, while the minimum reachable EO values (i.e., the ones favouring group $\agroup[1]$ (African-Americans)) are affected when increasing $\nexamples$, it is not the case for the maximum ones, as evidenced through consistent positive ranges.
Finally, we also observe that slightly tighter sparsity requirements can be met when increasing the training set size, although this significantly narrows the range of achievable fairness values.}

\begin{table}[h!]
\begin{center}
\begin{tabular}{cccc}
\hline
  $\nexamples$   &   Train accuracy   &   Test accuracy    &     Sparsity      \\
\hline
 500  & 0.720 ($\pm$0.012)  & 0.647 ($\pm$0.012) & 16.4 ($\pm$2.059) \\
 2000 & 0.690 ($\pm$0.005)  & 0.665 ($\pm$0.003) & 16.6 ($\pm$1.020)  \\
 3000 & 0.686 ($\pm$0.004) & 0.667 ($\pm$0.005) & 18.4 ($\pm$2.154) \\
 4000 & 0.684 ($\pm$0.003) & 0.672 ($\pm$0.004) & 19.6 ($\pm$1.356) \\
\hline
\end{tabular}
\end{center}
\caption{Training set accuracy, test set accuracy and sparsity (number of non-zero coefficients) of reference scoring systems on the COMPAS dataset, for different numbers of training examples $\nexamples$. We report both the average value and standard deviation.}\label{tab:ssperfs_scalability}
\end{table}

\newpage
\section{Complete Experimental Results}
\label{appendix:complete_results}

We hereafter report the results of all our experiments on scoring systems and decision diagrams across the three datasets (UCI Adult Income, Default of Credit Card Clients, and COMPAS) and the two fairness metrics (statistical parity and equal opportunity) considered. 

\updated{First, for each dataset, we report in Tables~\ref{tab:ss_perfs} (for scoring systems) and~\ref{tab:dd_perfs} (for decision diagrams) the predictive performances (training and test accuracies) and sparsities of the reference models (achieving the optimal loss \smash{$\emploss{\data}{\h_{\data}}$}). They confirm that the model's training accuracies are in line with the literature and that they generalize reasonably well.}

\begin{figure*}[!ht]
    \centering
    \begin{subfigure}{1\textwidth}
      \centering
       \includegraphics[width=0.48\linewidth]
        {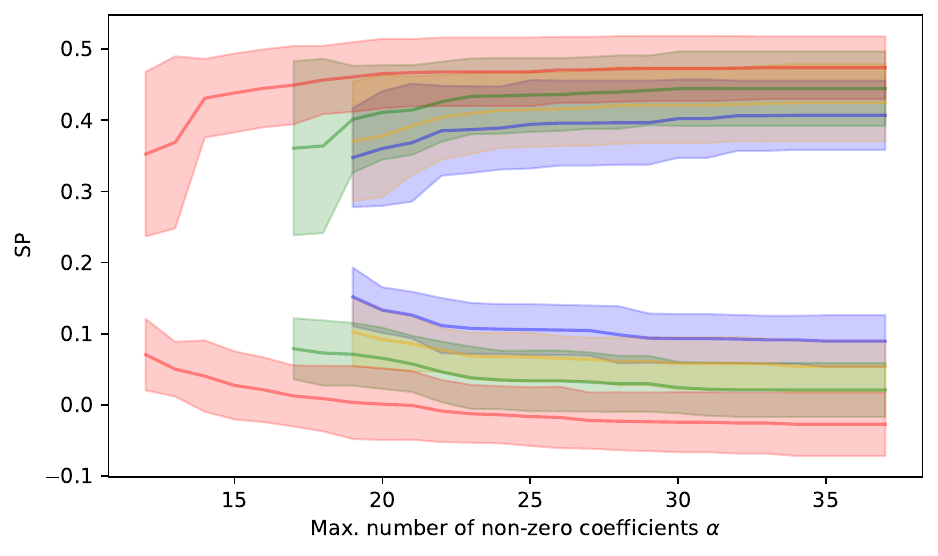}
        \hfill 
        \includegraphics[width=0.48\linewidth]
        {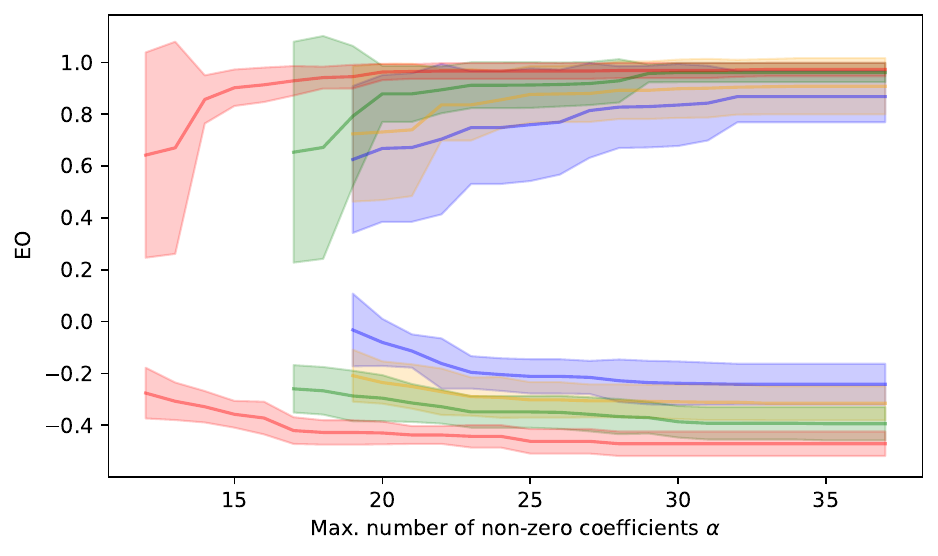}
      \caption{UCI Adult Income dataset. Left: statistical parity (SP), Right: equal opportunity (EO).}
      \label{fig:results_all_scoring_systems_adult}
    \end{subfigure}%

    \hspace{10pt}

    \begin{subfigure}{1\textwidth}
      \centering
       \includegraphics[width=0.48\linewidth]
        {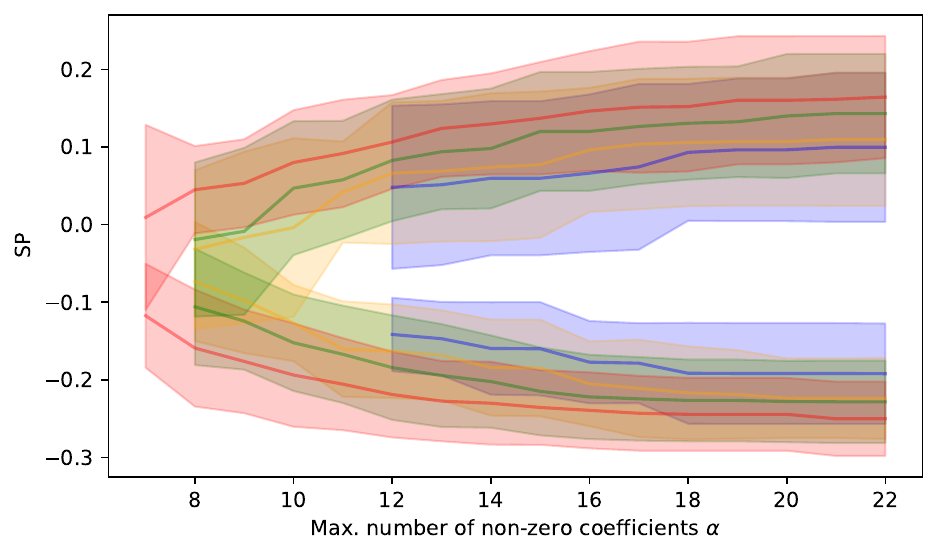}
        \hfill 
        \includegraphics[width=0.48\linewidth]
        {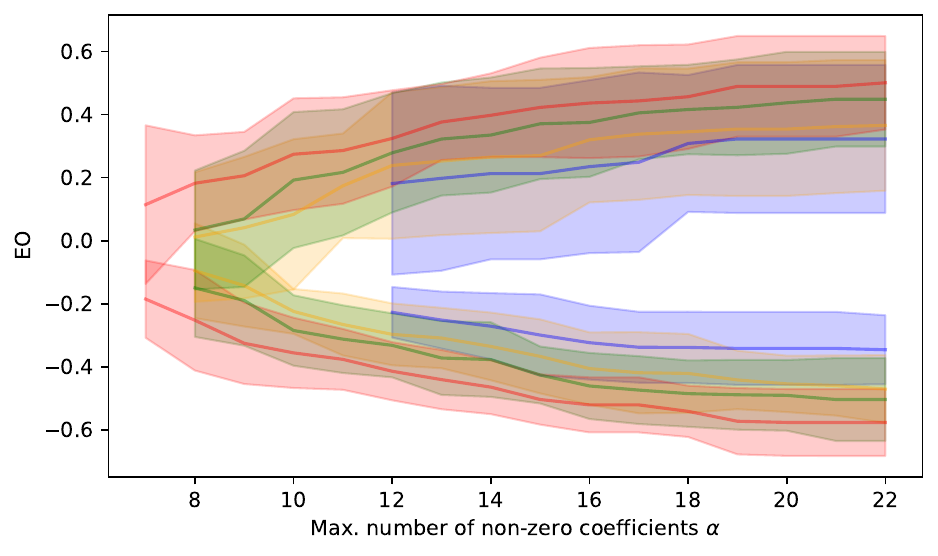}
      \caption{Default of Credit Card Clients dataset. Left: statistical parity (SP), Right: equal opportunity (EO).}
      \label{fig:results_all_scoring_systems_credit}
    \end{subfigure}%
    
    \hspace{10pt}
    
    \begin{subfigure}{1\textwidth}
      \centering
       \includegraphics[width=0.48\linewidth]
        {full_results_plot_ssystems_compas_SP_500_five-seeds_average.pdf}
        \hfill 
        \includegraphics[width=0.48\linewidth]
        {full_results_plot_ssystems_compas_EO_500_five-seeds_average.pdf}
      \caption{COMPAS dataset. Left: statistical parity (SP), Right: equal opportunity (EO).}
      \label{fig:results_all_scoring_systems_compas}
    \end{subfigure}%

    \hspace{10pt}

     \includegraphics[width=0.50\linewidth]{legend.pdf}

    \caption{Minimum and maximum statistical parity (SP) and equal opportunity (EO) achievable on the three considered datasets, within different 
    $\epsilon$-Rashomon sets for scoring systems, as a function of the chosen sparsity value $\sparsityvalue$ (number of non-zero coefficients in the built scoring systems). %
    We report both the average value (line) and standard deviation (colored area).}
\label{fig:results_all_scoring_systems}
\end{figure*}

\begin{figure*}[!ht]
    \centering
    \begin{subfigure}{1\textwidth}
      \centering
       \includegraphics[width=0.48\linewidth]
        {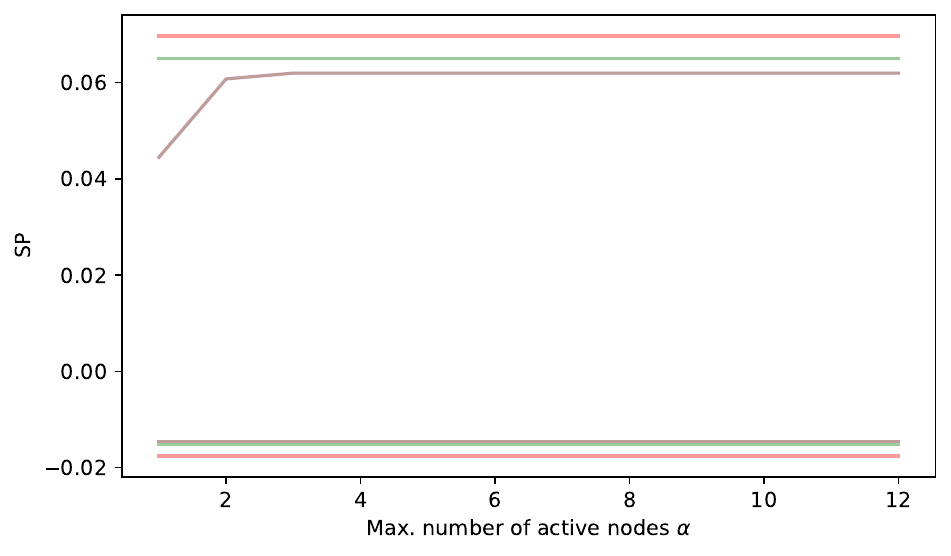}
        \hfill 
        \includegraphics[width=0.48\linewidth]
        {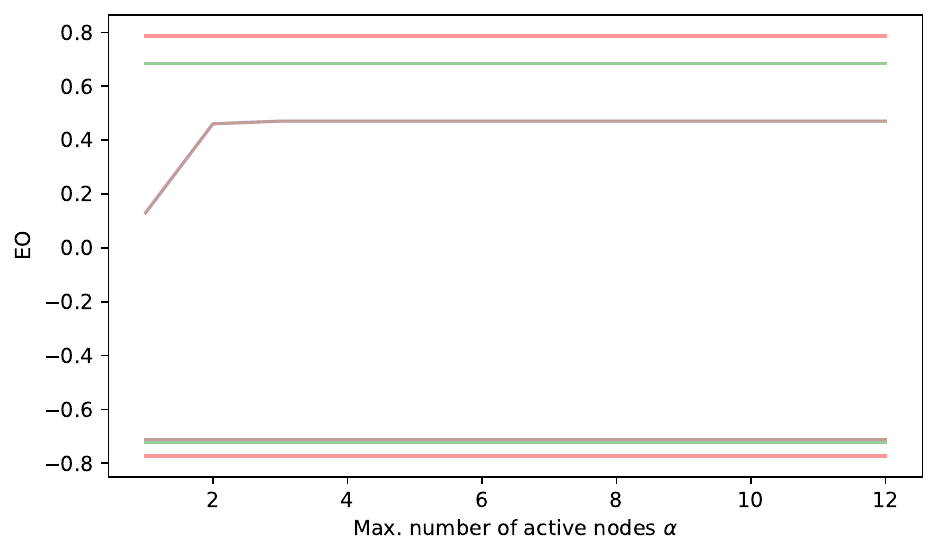}
      \caption{UCI Adult Income dataset. Left: statistical parity (SP), Right: equal opportunity (EO).}
      \label{fig:results_all_decision_diagrams_adult}
    \end{subfigure}%

    \hspace{10pt}

    \begin{subfigure}{1\textwidth}
      \centering
       \includegraphics[width=0.48\linewidth]
        {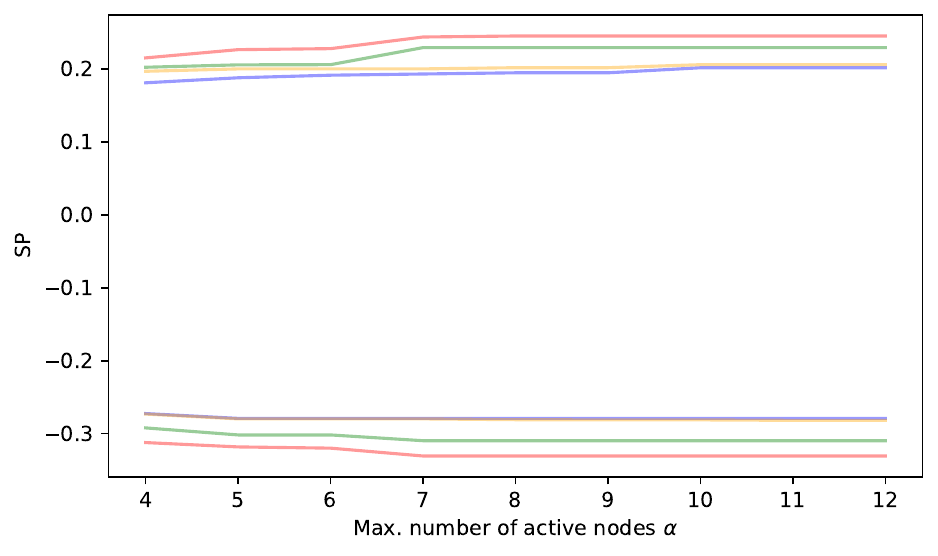}
        \hfill 
        \includegraphics[width=0.48\linewidth]
        {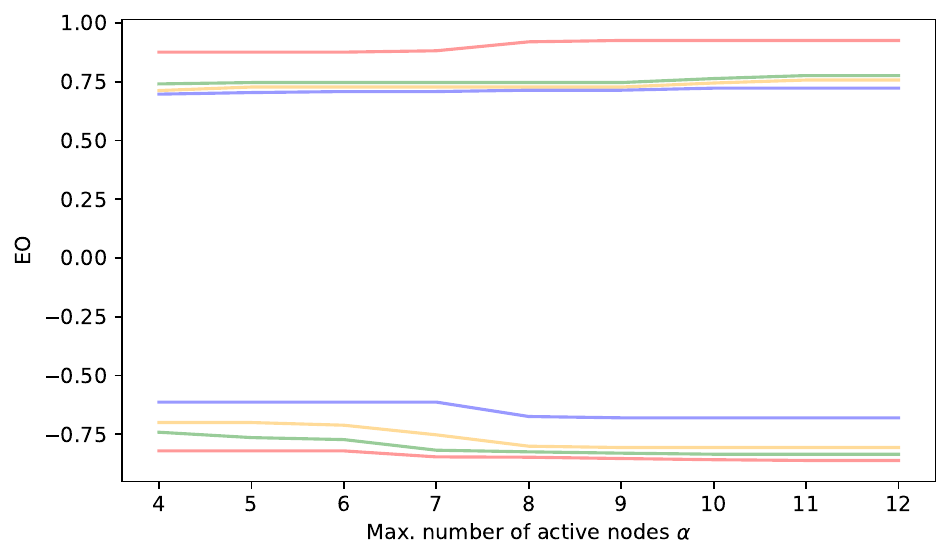}
      \caption{Default of Credit Card Clients dataset. Left: statistical parity (SP), Right: equal opportunity (EO).}
      \label{fig:results_all_decision_diagrams_credit}
    \end{subfigure}%
    
    \hspace{10pt}
    
    \begin{subfigure}{1\textwidth}
      \centering
       \includegraphics[width=0.48\linewidth]
        {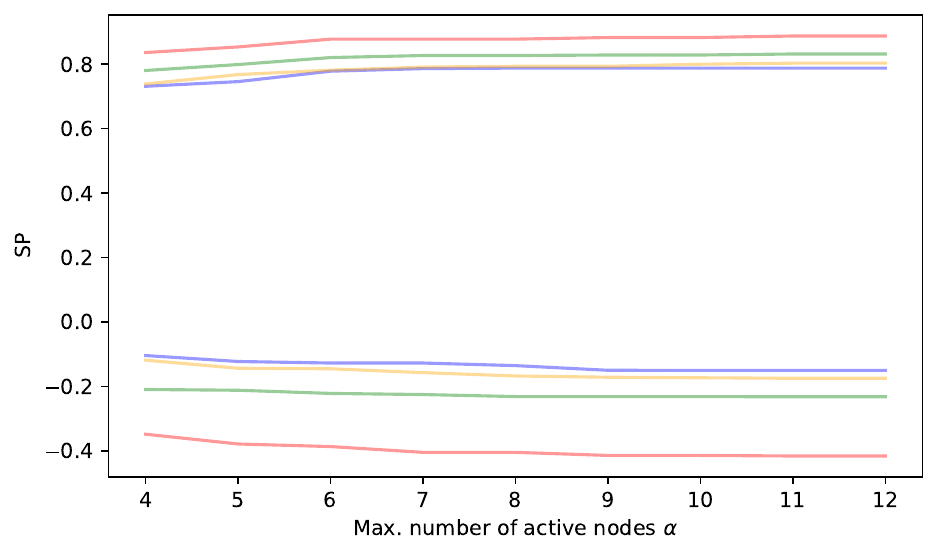}
        \hfill 
        \includegraphics[width=0.48\linewidth]
        {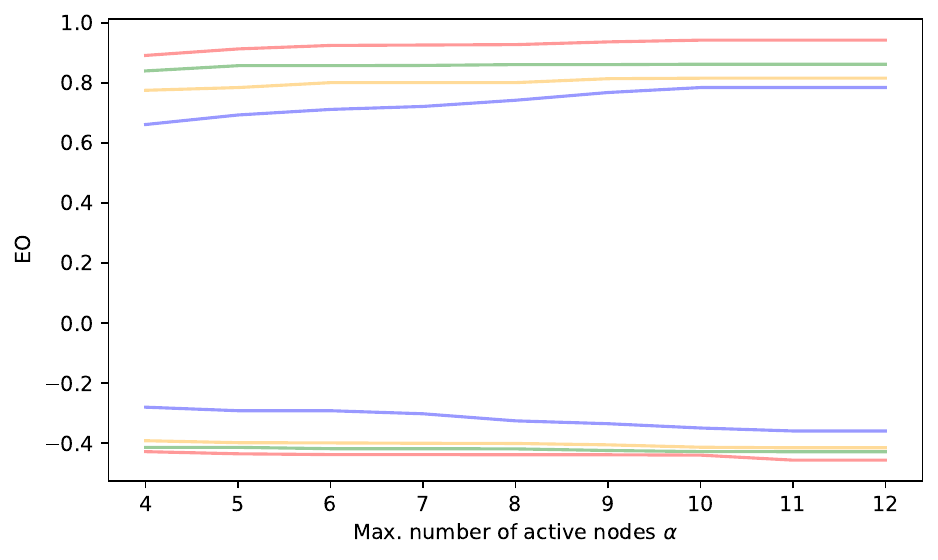}
      \caption{COMPAS dataset. Left: statistical parity (SP), Right: equal opportunity (EO).}
      \label{fig:results_all_decision_diagrams_compas}
    \end{subfigure}%

    \hspace{10pt}

     \includegraphics[width=0.50\linewidth]{legend.pdf}

    \caption{Minimum and maximum statistical parity (SP) and equal opportunity (EO) achievable on the three considered datasets, within different 
    $\epsilon$-Rashomon sets for decision diagrams, as a function of the chosen sparsity value $\sparsityvalue$ (number of active nodes in the built decision diagrams). %
    As the standard deviation areas significantly overlap, we only display the average value (line) and refer to the source code repository for the standard deviation values.}
\label{fig:results_all_decision_diagrams}
\end{figure*}

Second, for each dataset and fairness metric, we report one summary figure illustrating the complex interplays between the three considered desiderata: predictive performance, fairness, and sparsity. As in Figure~\ref{fig:effect3} (Section~\ref{sec:results}), we plot the minimum and maximum achievable fairness values as a function of the desired sparsity level $\sparsityvalue$ for different $\epsilon$ parameters. Figure~\ref{fig:results_all_scoring_systems} (respectively, Figure~\ref{fig:results_all_decision_diagrams}) reports all results for our experiments on scoring systems (respectively, on decision diagrams).
While all the experimental results can be read in these figures, detailed result files and tables are also available on our online repository\footnotemark[3], along with the source code. %

As discussed in Section~\ref{sec:results}, the greater complexity of decision diagrams using multivariate splits results in more nuanced trade-offs between predictive performance, fairness, and sparsity. More precisely, in Figure~\ref{fig:results_all_decision_diagrams}, the expected trends are observed: tightening either the Rashomon set parameter $\epsilon$ or the enforced sparsity value $\sparsityvalue$ further restricts the range of achievable fairness values. 
For instance, in our experiments using decision diagrams on the COMPAS dataset, for sparsity values $\sparsityvalue=4$, tightening the Rashomon set parameter $\epsilon$ from $20\%$ to $1\%$ restricts the minimum achievable statistical parity value from $-0.34$ to $-0.10$. On the same set of experiments, tightening the sparsity value $\sparsityvalue$ from $12$ to $4$ restricts the minimum achievable statistical parity within a $20\%$-Rashomon set from $0.41$ to $0.34$.
Although these differences are substantial, they remain subtler than those observed in our experiments on scoring systems. While the use of multivariate splits arguably affects the interpretability of the resulting models, this underscores the importance of the hypothesis class as a critical factor.

\begin{table*}[b]
\begin{center}
\begin{tabular}{cccc}
\hline
            Dataset             &   Train accuracy   &   Test accuracy    &     Sparsity      \\
\hline
        UCI Adult Income        & 0.884 ($\pm$0.012) & 0.81 ($\pm$0.005)  & 17.4 ($\pm$4.317) \\
 Default of Credit Card Clients & 0.842 ($\pm$0.018) & 0.791 ($\pm$0.005) & 12.6 ($\pm$2.332) \\
             COMPAS             & 0.720 ($\pm$0.012)  & 0.647 ($\pm$0.012) & 16.4 ($\pm$2.059) \\
\hline
\end{tabular}
\caption{Training set accuracy, test set accuracy and sparsity (number of non-zero coefficients) of reference scoring systems on the three considered datasets. We report both the average value and standard deviation.}\label{tab:ss_perfs}
\end{center}
\end{table*}

\begin{table*}[b]
\begin{center}
\begin{tabular}{cccc}
\hline
            Dataset             &   Train accuracy   &   Test accuracy    &     Sparsity     \\
\hline
        UCI Adult Income        & 0.916 ($\pm$0.024) & 0.776 ($\pm$0.029) & 10.4 ($\pm$0.8)  \\
 Default of Credit Card Clients & 0.847 ($\pm$0.022) & 0.774 ($\pm$0.01)  & 3.6 ($\pm$0.49)  \\
             COMPAS             & 0.638 ($\pm$0.091) & 0.58 ($\pm$0.045)  & 5.2 ($\pm$2.993) \\
\hline
\end{tabular}
\caption{Training set accuracy, test set accuracy and sparsity (number of active nodes) of reference decision diagrams on the three considered datasets. We report both the average value and standard deviation.}\label{tab:dd_perfs}
\end{center}
\end{table*}

\end{document}